\documentclass[preprint,12pt]{elsarticle}

\usepackage{amssymb}
\usepackage{color}
\usepackage{subcaption}

\usepackage{multirow}
\usepackage{amsmath}
\usepackage{csquotes}
\usepackage{algorithm}
\usepackage{algpseudocode}
\usepackage{tabularx}
\usepackage{caption}
\usepackage{subcaption}
\usepackage{adjustbox}
\usepackage{listings}
\usepackage{xcolor}
\usepackage{pifont}
\lstdefinestyle{json}{
    language=Java,
    showstringspaces=false,
    basicstyle=\ttfamily\small,
    keywordstyle=\color{blue},
    stringstyle=\color{blue},
    breaklines=true,
}

\usepackage{url}
\begin{document}

\begin{frontmatter}

\title{\textsc{Nbias}: A Natural Language Processing Framework for \textsc{bias} Identification in Text}

\author[inst1]{Shaina Raza\corref{cor1}}
\ead{shaina.raza@vectorinstitute.ai}

\author[inst2]{Muskan Garg}
\ead{muskanphd@gmail.com} 

\author[inst3]{Deepak John Reji}
\ead{deepak.reji@erm.com} 

\author[inst4]{Syed Raza Bashir}
\ead{syedraza.bashir@torontomu.ca} 

\author[inst4]{Chen Ding}
\ead{cding@torontomu.ca} 

\cortext[cor1]{Corresponding author}

\address[inst1]{Vector Institute for Artificial Intelligence, Toronto, ON, Canada}
\address[inst2]{Artificial Intelligence \& Informatics, Mayo Clinic, Rochester, MN, USA}
\address[inst3]{Environmental Resources Management, Bengaluru, Karnataka, India }
\address[inst4]{Toronto Metropolitan University, Toronto, ON, Canada}

\begin{abstract}
\noindent Bias in textual data can lead to skewed interpretations and outcomes when the data is used. These biases could perpetuate stereotypes, discrimination, or other forms of unfair treatment. An algorithm trained on biased data may end up making decisions that disproportionately impact a certain group of people. Therefore, it is crucial to detect and remove these biases to ensure the fair and ethical use of data. To this end, we develop a comprehensive and robust framework \textsc{Nbias} that consists of four main layers:  data, corpus construction, model development and an evaluation layer. The dataset is constructed by collecting diverse data from various domains, including social media, healthcare, and job hiring portals. As such, we applied a transformer-based token classification model that is able to identify bias words/ phrases through a unique named entity \textit{BIAS}. In the evaluation procedure, we incorporate a blend of quantitative and qualitative measures to gauge the effectiveness of our models. We achieve accuracy improvements ranging from 1\% to 8\% compared to baselines. We are also able to generate a robust understanding of the model functioning. The proposed approach is applicable to a variety of biases and contributes to the fair and ethical use of textual data.
\end{abstract}

\end{frontmatter}


\section{Introduction}
\label{sec:sample1}
\noindent The recent surge in Natural Language Processing (NLP) applications, encompassing fields from recommendation systems to social justice and employment screening, has sparked a critical concern - the emergence of bias within these systems \cite{hutchinson-etal-2020-social}. Instances of racial and gender bias have been increasingly reported \cite{bolukbasi2016man}, indicating an urgent need for scrutiny. These biases often originate from the training data used in NLP models, and a majority of these large datasets harbor inherent biases . Regrettably, many NLP practitioners lack the necessary awareness or knowledge to effectively identify and address these biases, highlighting a significant gap in the field. 

Furthermore, there is a notable lack of discussion on data specifics - its origin, generation, and pre-processing - in many NLP publications. Given these circumstances, the importance of addressing biases in NLP applications cannot be overstated. These biases, if unchecked, not only compromise the validity of the models, but can also have unfavorable and detrimental consequences. The objective of this research is to provide insights into the detection of bias in NLP datasets, contributing to the development of more equitable and unbiased Artificial Intelligence (AI) systems.

Bias in text data is a pervasive and deeply-rooted issue. The bias in data often stems from cognitive predispositions that influences our dialogues, views, and understanding of information \cite{dixon2018measuring}. This bias can be explicit  which are often seen in discriminatory language targeting certain racial or ethnic groups \cite{ribeiro2018media}, as in social media. Implicit bias \cite{yanbo2020implicit}, on the other hand, subtly perpetuates prejudice through unintentional language use but is equally harmful.

\begin{figure}
    \centering
    \includegraphics[width=1\linewidth]{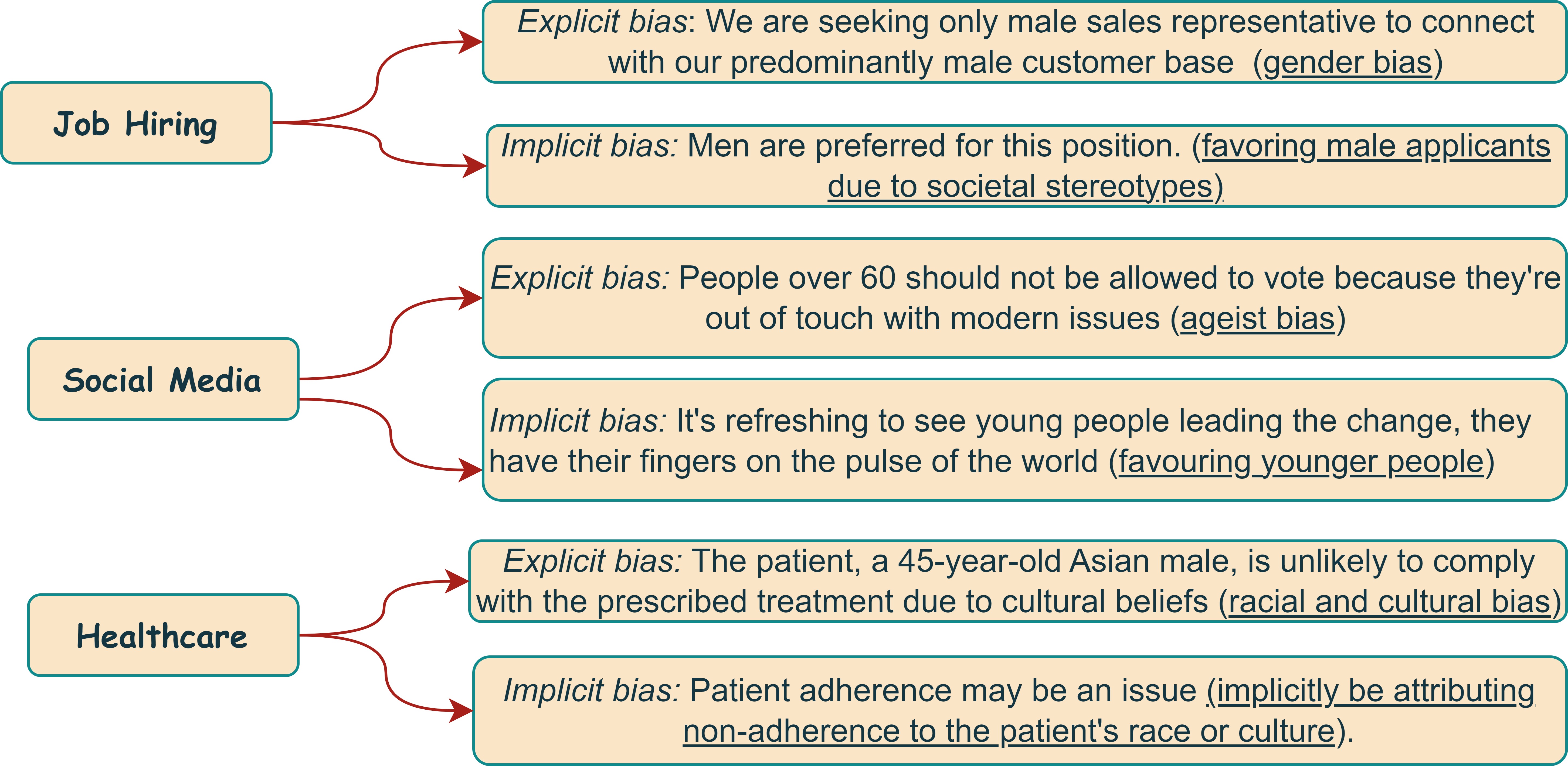}
   \caption{Visual Representation of Implicit and Explicit Biases in Textual Data: Examples from Job Hiring, Social Media, and Healthcare.}
 
    \label{fig:1}
\end{figure}

The necessity for unbiased, trustworthy text data has grown across sectors like healthcare \cite{Thomasian2021}, social media \cite{ribeiro2018media,raza2022dbias}, and recruitment \cite{gaucher2011evidence}. This data is essential for training NLP models for various downstream tasks, like formulating healthcare diagnoses and treatment plans, handling discriminatory language on social media, and promoting fair recruitment practices. Figure \ref{fig:1} illustrates the complexities of biases in text data in various domains, including job hiring, social media, and healthcare. These biases are primarily conveyed through lexical choices \cite{dawkins-2021-marked} and demand sophisticated detection methods, motivating this research. The primary aim of this study is to further foundational research on the fairness and reliability of the textual data. 

\noindent Although NLP has advanced much, the state-of-the-art techniques \cite{bolukbasi2016man,spinde-etal-2021-neural-media,Farber2020} often concentrate on bias detection in specific domains and lack generalizability. To address this, our research offers a generalizable bias detection method proven effective across the various domains. We present \textsc{Nbias}, a comprehensive framework for detecting bias in text data. This involves data preparation where bias-indicative terms are marked using a transformer-based token classification method like Named Entity Recognition (NER).  

Current NER solutions can manage general \cite{nie-etal-2020-named}, biomedical \cite{Raza2023}, and social media \cite{moon-etal-2018-multimodal} entities, but often neglect \textit{BIAS} as a separate entity. To address this, we introduce a new entity type, \textit{BIAS}, to identify biased terms in text data. In this context, bias refers to unfair and often harmful favoritism or prejudice towards a particular group, person, or idea, which can manifest through profanity, unjustified criticism, or discriminatory language.

A key contribution of this study is the development of the first comprehensive framework for bias detection in text data. This framework is based on latest language model technology and incorporates four crucial layers: data gathering, corpus construction, model development, and rigorous evaluation. The specific contributions of the work are as follows:

\begin{enumerate}
    \item \textit{Development of Annotated Datasets}: Acknowledging the scarcity of bias annotations in text-based data, we designed a solution by generating multiple annotated datasets. Our work fills a critical gap in the available resources, thereby providing a solid foundation for future research in the realm of bias detection.
    \item \textit{Semi-Autonomous Labeling}: To alleviate the time-intensive manual annotation process, we pioneered a novel methodology termed \enquote{semi-autonomous labeling}. This strategy provides a faster and more efficient way of annotating bias-related terms within textual content. This innovative approach has significant implications for improving the speed and accuracy of bias detection.
    \item \textit{Unique Entity Type - BIAS}: In an effort to enhance the precision of bias identification within text, we introduced a unique entity type, \textit{BIAS}. This new entity type is specifically designed for detecting biased words and phrases within the text data. This has the potential to dramatically improve the process of bias identification and quantification in text-based analysis.
    \item \textit{Comprehensive Evaluation Process}: We subjected our proposed framework to a thorough evaluation process, utilizing both quantitative and qualitative analysis methods. The results confirm the reliability and efficiency of our approach, making it compatible for its application in real-world scenarios. This rigorous evaluation sets a benchmark for assessing the efficacy of bias detection methodologies.
\end{enumerate}

\section{Related Work}
\subsection{Identifying Bias in NLP}
\noindent One of the key challenges associated with NLP systems lies in the presence of bias, a manifestation of unfair and systematic discrimination observed in their outcomes \cite{garrido2021survey}. Moreover, the past studies \cite{caliskan2017semantics, spinde-etal-2021-neural-media,Farber2020, bolukbasi2016man, dev2021measures} have shown the societal and cultural prejudices are deeply embedded within the training data due to the presence of bias. As such, the biases, whether explicit or implicit, can significantly impact the functionality of the NLP systems, leading to skewed results and perpetuating existing societal biases. Thus, the detection and mitigation of these biases are crucial to promoting fairness and inclusiveness within NLP systems \cite{raza2022dbias,Farber2020}.

Researchers have proposed and implemented various strategies to identify bias, including employing statistical methods to discover patterns of bias within the training data \cite{bolukbasi2016man,Manzini2019}. Under this approach, specific words or phrases that appear to be disproportionately associated with certain demographic groups, such as genders or races, are identified. For example, certain adjectives might be used more frequently in descriptions of women than men \cite{bolukbasi2016man}, or vice versa. The identification and debiasing of such patterns can highlight areas of potential bias, providing a starting point for efforts to eliminate these biases \cite{Tokpo2023}.

The field of bias detection in NLP has seen a surge of innovative methods in recent years, primarily leveraging advanced machine learning techniques. One such study has considered the combination of a speech detection system with an explanatory method to identify potential bias \cite{cai2022power}. In this method, not only is the system trained to detect instances of hate speech, but it also provides a rationale or explanation for its classification decisions. Another area of research that has attracted considerable attention is the investigation of bias in event detection datasets and models \cite{wang2023m4}. Event detection tasks involve identifying and classifying real-world events within text data. These tasks can be susceptible to a range of bias-related issues, including data sparsity, labeling task, and annotations.

Additionally, NLP techniques have been employed to address various aspects of bias. For instance, in a related study \cite{pair2021quantification} on gender bias and sentiment towards political leaders in the news were quantified using word embeddings and sentiment analysis. Another work focused on investigating ableist bias in NLP systems, particularly at the intersection of gender, race, and disability \cite{hassan-etal-2021-unpacking-interdependent}. Similarly, a methodology was proposed to eliminate gender bias from word embeddings \cite{ding2022word}. Furthermore, marked attribute bias in natural language inference was identified and analyzed, with an evaluation of existing methods for bias mitigation \cite{dawkins-2021-marked}. These studies provide a deep understanding of the social and cultural factors that contribute to bias identification.

Another work \cite{govindarajan2023people} presents bias analysis in NLP beyond demographic bias, focusing on predicting interpersonal group relationships using fine-grained emotions. A related study \cite{devinney2022theories} evaluates gender bias in NLP research, highlighting the lack of explicit gender theorization. In another work, authors \cite{zhao2023combating} introduce an effective bias-conflicting scoring method and gradient alignment strategy to identify and mitigate dataset biases. Overall, these studies underscore the importance of continuous efforts in identifying and mitigating biases in models to ensure fairness and equity.

\subsection{Named Entity Recognition (NER)}
\noindent Named Entity Recognition (NER) is a significant task in NLP that is aimed at identifying and classifying named entities within textual data. NER is a token classification task that focuses on identifying and classifying named entities such as individuals, organizations, and locations within a given text. In the past, many traditional methods have been employed for NER, each with its unique characteristics and benefits.

\begin{itemize}
    \item Rule-based methods rely on predefined sets of rules to identify named entities \cite{eftimov2017rule}. This method usually employs regular expressions or dictionary-based techniques to extract entities. Although rule-based methods can be effective for well-defined and specific contexts, their performance can decrease in the face of variability and ambiguity in language usage.

    \item Supervised learning methods leverage annotated data to train a model for NER \cite{moon-etal-2018-multimodal,chiu2016named}. These methods use statistical models such as Support Vector Machines (SVM), Conditional Random Fields (CRF), and others to classify the named entities. The performance of supervised learning methods can be impressive, given sufficient high-quality annotated data.

    \item Deep learning methods, which are more contemporary approaches, utilize complex architectures like recurrent neural networks (RNNs) and transformer-based language models to extract named entities \cite{liu2021tner,Raza2023}. These methods have shown promising results in NER tasks, owing to their capacity to capture intricate language patterns and contextual information.

\end{itemize}

\noindent A recent study introduced a contrastive learning-based approach for multimodal NER \cite{liu2023reducing}. This approach leverages both textual and non-textual data to identify and classify named entities, harnessing the complementary information offered by different modalities to improve the model's performance. Another research work into event detection from social media posts, evaluating the effectiveness of a pre-trained NER model followed by graph-based spectral clustering \cite{liu2022social}. The study also explored transformer-based methods to weight the edges of the graph for event detection, further refining the detection process. A span-based NER model eliminates the need for label dependency \cite{liu2022social}. This approach addresses the issue of cascade label mis-classifications, a common challenge in traditional NER models that depend on label sequences.

While our work on token classification is inspired by these studies, we identify a notable gap in the literature: the existing seminal work does not recognize \textit{BIAS} as an entity. In this work, we detect biased expressions within unstructured texts, designating them under the 'BIAS' entity label.

\subsection{Data Annotation}
\noindent Data annotation is a crucial task in NLP as it involves labeling and categorizing information to extract valuable insights from text data \cite{Raza2023a}. By enriching text data with relevant metadata, such as part-of-speech tags, named entity tags, and sentiment tags, data annotation provides contextual information that is essential for subsequent analysis \cite{gerstenberger2017instant}. Quality annotated data enhances model learning, boosting prediction accuracy. In contrast, inadequate annotations impede learning, resulting in subpar performance. Various methods of data annotation cater to different requirements of speed, quality, and computational resources:

\begin{itemize}
    \item Manual annotation is carried out by human annotators who carefully review and label the data. This method typically yields high-quality results, given the nuanced understanding that humans have of language. However, manual annotation is often time-consuming and labor-intensive, and its feasibility may be  limited by the availability of qualified annotators and financial resources \cite{eftimov2017rule}.

    \item Semi-automatic annotation combines manual efforts with automated tools to accelerate the annotation process and minimize human error. These tools can range from rule-based systems to pre-trained machine learning models \cite{rebuffi2020semi}. While semi-automatic annotation can improve efficiency, its accuracy may still depend on the quality of the automated tools and the manual review process.

    \item Automatic annotation leverages machine learning models and algorithms to annotate text data without human intervention \cite{alex2010agile}. Although automatic annotation can process vast amounts of data in a relatively short time, its accuracy may be compromised, particularly for complex or ambiguous texts. Therefore, a common practice is to combine automatic annotation with manual review to ensure data quality.
\end{itemize}

\noindent Various strategies have been developed to address these challenges and optimize the annotation process. One study presents a comprehensive comparison of different annotation tools, highlighting their strengths and limitations \cite{Caufield2019}. Another research work proposes a method for automatically generating high-quality labeled data for NER tasks by leveraging existing knowledge bases \cite{fieldmatters-2023-nlp}. A similar study has developed an annotation framework that combines statistical machine translation and human annotation to create a parallel corpus \cite{ghaffari2022adversarial}. Other researchers have investigated methods for improving the reliability and consistency of manual annotations, such as developing guidelines and protocols for annotation tasks \cite{green2018proposed}or implementing quality control mechanisms to ensure data quality.

Ultimately, the choice of annotation method and tools will depend on the specific requirements of a project, such as the desired level of accuracy, the available resources, and the nature of the data being annotated.To this end, we employ a \textit{semi-automatic annotation} strategy, integrating human proficiency with semi-supervised learning methodologies.

\section{Proposed Framework for Bias Identification in Texts}
\begin{figure}
    \centering
    \includegraphics[width=1\linewidth]{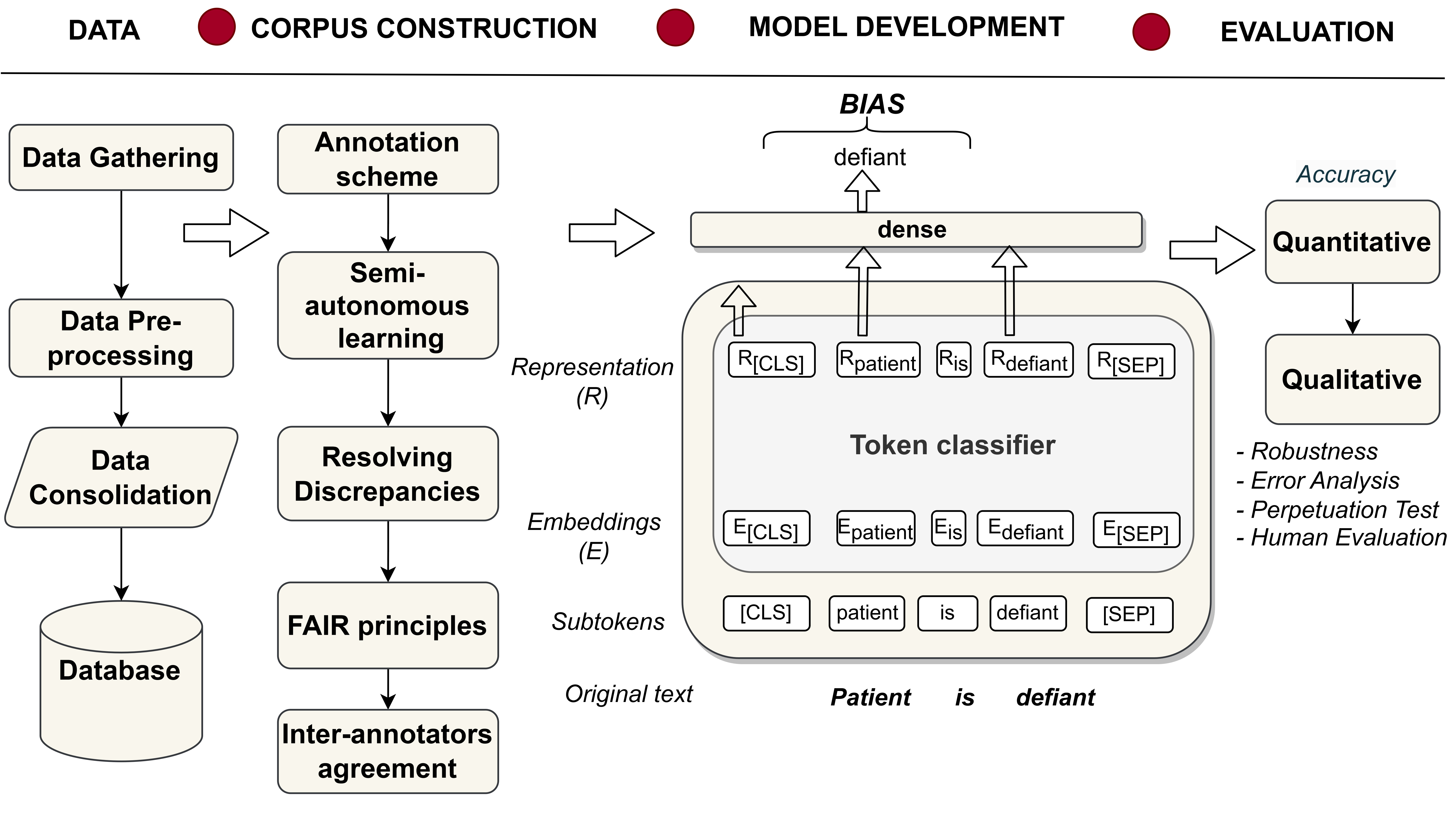}
    \caption{\textsc{Nbias}: A Natural Language Processing Framework for Bias Identification.}
    \label{fig:2}
\end{figure}

\noindent In this section, we present \textsc{Nbias}, an innovative framework designed to detect biases within textual data, as illustrated in Figure \ref{fig:2}.  The \textsc{Nbias} framework is structured into four distinct layers: (i) the data collection layer, (ii) the corpus construction layer, (iii) the model development layer, and (iv) the evaluation Layer. Each layer is designed to collaborate seamlessly with the others, providing an effective and comprehensive approach for detecting biases in textual content.

\subsection{Data Layer}
\noindent The Data Layer serves as the framework's primary interface with the data for analysis. It handles data collection, pre-processing and data consolidation from a variety of sources, such as social media, online articles, and databases. This layer ensure adaptability and high performance for the entire framework.

\paragraph{Data Gathering} 
Our study adopts a methodological data collection approach, incorporating diverse sources from various domains. To analyze biases in medical narratives and healthcare, we include data from two important clinical text databases: MIMIC-III \cite{MIMICIII75:online} and MACCROBAT \cite{Caufield2019}. The MIMIC-III dataset is a publicly available database with de-identified health data from over 40,000 ICU patients. It offers rich clinical narratives, including nursing notes, radiology reports, and discharge summaries, enabling a deep understanding of biases in healthcare communication. The textual data were primarily obtained from the \textit{NOTEEVENTS} table.

The MACCROBAT dataset provides valuable pediatric critical care data, including admission notes and medical summaries. It contains 200 original documents along with corresponding annotated versions centered around clinical case reports.

To detect bias in news articles and social media streams, we use the BABE (Bias Annotations By Experts) dataset \cite{spinde-etal-2021-neural-media}. This dataset includes 3700 articles and tweets, offering a comprehensive perspective on linguistic bias in media and public opinion. It features marked statements, enabling recognition of bias at both granular (word-level) and broader (sentence-level) scopes, covering diverse topics.

To examine biases in employment practices, we incorporate the Job Hiring/Recruitment dataset \cite{Classify68:online}, comprising 20,000 job posts with titles, descriptions, and associated tags from various businesses. Each advertisement includes job details and manually assigned tags by recruiters, suggesting jobs to potential candidates with analogous skills.

\paragraph{Data Consolidation} 
After gathering and pre-processing data from various sources, all datasets are harmonized into a single consolidated dataframe. This dataframe includes the following columns:

\begin{itemize}
    \item \textbf{Dataset}: Specifies the source dataset, such as MIMIC-III, MACCROBAT, Job Hiring, or BABE
    
    \item \textbf{Text}: Contains the actual textual data extracted from the respective datasets, including clinical notes, case reports, job descriptions, or annotated statements.
    
    \item \textbf{Biased Words}: Includes the words or phrases identified as biased in the text, crucial for granular bias detection.
    
    \item \textbf{Aspect of Bias}: Denotes the specific type or aspect of bias present in the text, categorized by gender, racial, or age biases, to understand the nature of the biases detected.
    
    \item \textbf{Label}: Indicates whether the text is biased or non-biased, serving as the target variable for the token classifier and for evaluation purposes.
\end{itemize}

A sample record in JSON format is shown below:
\begin{lstlisting}[style=json]
{
  "Record": {
    "Dataset": "MIMIC-III",
    "Text": "Clinical notes of patient XYZ indicate a history of superficial hypertension due to overly emotional personality.",
    "BiasedWords": "superficial, overly emotional personality",
    "AspectOfBias": "age",
    "Label": "biased"
  }
}
\end{lstlisting}

\noindent In the consolidated dataframe, each row represents a unique sample from the original dataset, supplying information for bias detection and assessment. Further pre-processing is conducted to prepare the data for subsequent layers of the \textsc{Nbias} framework, particularly the NLP model performing token classification.

\paragraph{Data Pre-processing} 
The pre-processing of textual data involves a series of sequential operations to refine and structure the data for machine learning algorithms. This includes tokenization, which involves breaking raw text into meaningful tokens (words or subwords) for semantic understanding and subsequent NLP tasks; text cleaning, which involves removing punctuation, numbers, special characters, and converting text to lowercase to ensure uniformity and clarity; and handling missing values, which involves identifying and appropriately managing missing data to avoid bias and improve model performance. These pre-processing steps convert raw text into a clean, structured format, enabling the NLP token classification model in the subsequent layer.

\subsection{Corpus Construction}
\noindent Our group, consisting of three seasoned professionals from the disciplines of healthcare, computer science, and journalism, was joined by two diligent students to perform the task of detecting and labeling bias in our dataset. Their collective role centered around the critical task of carefully annotating bias within our dataset. This endeavor is important to ensure the integrity and fairness of any subsequent analysis or research. The foundation for this task was based on comprehensive guidelines that clearly delineated the concept of bias in this context. 

Bias, as per the instructions, was defined as any terminology or phraseology that could potentially provoke prejudiced comprehension or induce stereotyping, as mentioned in most of the literature \cite{Farber2020, raza2022dbias, spinde-etal-2021-neural-media} also. The factors from which biases could stem were identified as gender, race, socioeconomic status, age, or disability for this NLP work. Such biases could inadvertently skew the dataset and, consequently, the results derived from it. Thus, the identification and annotation of such biases are of high importance to uphold the accuracy and reliability of our dataset. Highlighting both explicit and implicit biases was emphasized as a critical part of our work.

\paragraph{Annotation Scheme}
Under the light of these provided guidelines, our team proceeded by using a carefully compiled list of terms and phrases, known as \enquote{bias-indicative} lexicons. These lexicons provided a comprehensive guide to potential areas where bias could lurk within our dataset. A portion of this list is exhibited in Table \ref{tab:1} for reference. This bias-indicative lexicon served as a navigational tool for our team to identify and mark \enquote{BIAS} entities scattered within our textual data. These entities can be individual words or phrases that express or imply bias. This systematic approach ensured that we could account for most biases that exists in the data.

\begin{table}[h]
\centering
\caption{Bias Dimensions and Fewer Sample Biased Words/Phrases Shown Due to Brevity}
\label{tab:1}
\small 
\begin{tabularx}{\textwidth}{|l|X|}
\hline
\textbf{Bias Dimension} & \textbf{Biased Words/Phrases} \\
\hline
Gender & ‘hysterical’, ‘emotional’, ‘weak’, ‘bossy’, ‘fragile’, ‘nagging’, ‘man up’, ‘tomboy’ \\
\hline
Race & ‘inner city’, ‘illegal alien’, ‘thug’, ‘exotic’, ‘uncivilized’, ‘model minority’, ‘white trash’ \\
\hline
Social Status & ‘trailer park’, ‘lazy’, ‘freeloader’, ‘welfare queen’, ‘ghetto’, ‘lazy bum’, ‘filthy rich’ \\
\hline
Age & ‘senile’, ‘slow’, ‘old-fashioned’, ‘whippersnapper’, ‘elderly’, ‘young and naive’, ‘generation gap’ \\
\hline
Disability & ‘handicapped’, ‘crippled’, ‘invalid’, ‘sufferer’, ‘differently-abled’, ‘victim’ \\
\hline
Religion & ‘radical’, ‘terrorist’, ‘infidel’, ‘heathen’, ‘fanatic’, ‘holy roller’ \\
\hline
Profession & ‘greedy’, ‘dishonest’, ‘corrupt politician’, ‘crooked lawyer’, ‘greedy CEO’, ‘lazy government worker’ \\
\hline
National & ‘unpatriotic’, ‘alien’, ‘foreigner’, ‘outsider’, ‘immigrant’, ‘nationalist’ \\
\hline
Education & ‘uneducated’, ‘illiterate’, ‘dropout’, ‘underachiever’, ‘overachiever’, ‘smarty-pants’ \\
\hline
Body Size & ‘fat’, ‘slob’, ‘skinny’, ‘lardass’, ‘beanpole’, ‘plus-sized’ \\
\hline
\end{tabularx}
\end{table}

\noindent We adopted the Inside-Outside-Beginning (IOB) annotation scheme \cite{Sexton2022} to classify and annotate ‘BIAS’ entities. This technique categorizes tokens in the text as the beginning (B), inside (I), or outside (O) of a bias entity. ‘B’ represents the first token of a bias entity, ‘I’ for tokens inside the entity, and ‘O’ for tokens not part of any bias entity. This approach ensured consistent and precise annotations, enhancing the reliability and accuracy of our study.

 \paragraph{Annotation Approach} We leveraged semi-supervised learning methodologies \cite{rebuffi2020semi,Raza2023,Raza2023a} to enhance both efficiency and precision of the annotation process. The integration of BERT (Bidirectional Encoder Representations from Transformers) , known for its superior text comprehension abilities, substantially improved our approach.

Our annotation process initiated with the manual tagging of 20\% of the complete dataset. This critical yet time-consuming task was strategically limited to a subset of data, ensuring a balance between accuracy and efficiency. The \enquote{BIAS} entities were carefully annotated in compliance with our predefined guidelines. This annotated subset then fed into our BERT model, serving as training data for the token-classification task.  Once sufficiently trained, the model was assigned the task of predicting \enquote{BIAS} entities within the remaining 80\% of the data. The extensive dataset was effectively managed by breaking it down into 20\% increments, a process we refer to as \enquote{semi-autonomous labelling}. Expert reviews cross-verified the \enquote{BIAS} entities labelled by the model. This combination of semi-supervised learning with expert validation enabled us to create an annotation process that is both optimized and trustworthy.

\paragraph{Working Instance}  To demonstrate our annotation scheme, we consider the example sentence:\textit{ \enquote{The overpriced product from the highly successful company was surprisingly popular}.} Table \ref{tab:2} presents the corresponding BIO format annotations for this sentence. Assuming the term \enquote{overpriced} holds potential bias, it would be tagged as \enquote{B} in the BIO scheme, indicating the start of a bias entity. All other tokens not part of this bias entity would be labeled \enquote{O}. This example shows our extensive annotation process across our dataset. This approach allows us to quantify and comprehend biases in a consistent manner.

\begin{table}[h!]
\centering
\caption{Bias Annotation using BIO scheme}
\label{tab:2}
\small 
\begin{tabular}{|c|c|}
\hline
\textbf{Word} & \textbf{Bias Annotation} \\ \hline
The & O \\ \hline
overpriced & B-BIAS \\ \hline
product & O \\ \hline
from & O \\ \hline
the & O \\ \hline
highly & B-BIAS \\ \hline
successful & I-BIAS \\ \hline
company & O \\ \hline
was & O \\ \hline
surprisingly & B-BIAS \\ \hline
popular & I-BIAS \\ \hline
. & O \\ \hline
\end{tabular}

\end{table}

\paragraph{Resolving Discrepancies} An integral part of our process was addressing discrepancies between annotators, a common criteria in multi-person annotation tasks. We implemented a consensus-driven approach to uphold consistency and reliability in our annotations. Any disagreement was discussed collectively, considering each annotator's viewpoint and reaching a unified decision based on predefined annotation guidelines. This process ensured collective agreement on all annotations, minimizing potential bias or error and boosting reliability. This consensus strategy was uniformly applied across all data sources including the BABE, MIMIC, MACCROBAT, or Job Hiring datasets. 

\paragraph{FAIR Principles} After reaching consensus on all annotations, we saved the final annotated data in the widely-accepted CoNLL-2003 format \cite{Spinde2021}. This format represents data in a tab-separated manner, associating each word with its part of speech tag, chunk tag, and named entity tag. Sentences are separated by an empty line, and each row corresponds to a token with its annotation.

The CoNLL-2003 format offers multiple benefits. It ensures compatibility with existing NLP tools and models, facilitating future analysis and model training. Additionally, it promotes collaboration and peer review by allowing easy data sharing and comprehension among researchers. Lastly, it enhances the reproducibility of our study, enabling others to use our data for model validation and findings replication. By adhering to the FAIR principles, our dataset is made \textbf{F}indable, \textbf{A}ccountable, \textbf{I}nteroperable, and \textbf{R}eusable, enhancing the transparency, accessibility, and reliability of our research.

\paragraph{Inter-Annotator Agreement} In our research, we placed considerable emphasis on establishing rigorous protocols to guarantee the reliability and consistency of the data annotations.  Two independent reviewers were assigned to carefully assess the annotated data, promoting objective evaluation devoid of influence from initial annotators. Rather than relying on subjective judgment, we quantified their agreement through Cohen's Kappa coefficient—a statistical measure common in categorical data studies, accounting for potential chance agreement. Scores over 0.6 denote \enquote{substantial} agreement and above 0.8 represent \enquote{almost perfect} agreement. Our reviewers attained a Cohen's Kappa score of 78\%, demonstrating high concordance on the annotations. This high score substantiates the uniformity, consistency, and quality of our annotations. Moreover, it demonstrates the objectivity of the assessment process, highlighting the well-built nature of our annotated data. This, in turn, enhances the trustworthiness of prospective findings drawn from this dataset.

\subsection{Model Development Layer}
\noindent In this layer, we leverage the BERT language model for token-classification and adapt it for the task of NER . The choice of BERT is motivated by its powerful capability of understanding both left and right context of a word, and its effectiveness in recognizing and classifying multi-word phrases. These features make it particularly well-suited for the complex task of bias detection and annotation in our text data.

The advantage of using BERT in  \textsc{Nbias} model development lies in its more effective token-level bias identification. \textsc{Nbias} incorporates enhancements to the standard BERT architecture, such as modifications in the attention mechanism, loss function, and fine-tuning approaches, specifically tailored for better capturing biases in complex text data. The subsequent section provides a detailed explanation of the model development.

The token classifier architecture (shown in as the middle component in Figure 2) consists of a multi-layer bidirectional transformer encoder that captures contextual information from both directions. Given an input sequence $X=\{x_1,x_2,...,x_n\}$, the words are tokenized and embedded as shown in Equation (1):
\begin{equation}
E(X) = \{e(x_1), e(x_2), \ldots, e(x_n)\} 
\end{equation}

\noindent where $E(X)$ represents the set of embedded representations for an input sequence $X$, $X$ consists of $n$ words $\{x_1,x_2,...,x_n\}$, $e(x_i)$ is the embedding function that maps each word $x_i$ from the input sequence to a continuous vector representation. The embedded input sequence is then passed through the transformer layers.

BERT employs self-attention mechanisms to weigh the importance of different words in the input sequence, enabling it to better identify and understand complex relationships between words. The self-attention $Att$ score between word $i$ and word $j$ is computed as shown in Equation (2):
\begin{equation}
Att(i,j) = \text{Softmax}\left(\frac{Q(e(x_i)) \cdot K(e(x_j))^T}{\sqrt{d_k}}\right)
\end{equation}

\noindent where $Q$, $K$ are the query and key matrices, and $d_k$ is the key dimension.

Following the transformer encoder, the output after applying self-attention and passing through the bidirectional transformer encoder is represented, as shown in Equation (3):
\begin{equation}
R(X)=\{r(x_1),r(x_2),...,r(x_n)\} 
\end{equation}
\noindent where $R(X)$ represents the set of contextualized representations for an input sequence $X$. The function $r(x_i)$ is the representation function that maps each word $x_i$ from the input sequence to a continuous vector representation after passing through the transformer encoder.

A linear layer with a softmax activation function is added for entity classification. This layer transforms the representations generated by the transformer encoder into a probability distribution over the possible output classes. To simplify our annotation and prediction task, we have merged the 'B' (Beginning) and 'I' (Inside) tags from the standard BIO tagging scheme into a single 'BIAS' tag. The 'BIAS' tag represent any part of a bias entity, while 'O' represents non-entity. The probability distribution is calculated as shown in Equation (4):
\begin{equation}
P(y|x) = \text{Softmax}(W \cdot c(x) + b) 
\end{equation}
\noindent where $W$ is the weight matrix, $b$ is the bias vector in the softmax function, and $P(y|x)$ is the probability distribution over the output classes `BIAS' and `O'.  The final output of the model indicates the presence of biased words or phrases within the input sequence by labeling them as `BIAS'. This simplification enables our model to recognize biased phrases more effectively, without differentiating between their start or continuation.

\begin{figure}[H]
    \centering
    \includegraphics[width=0.75\linewidth]{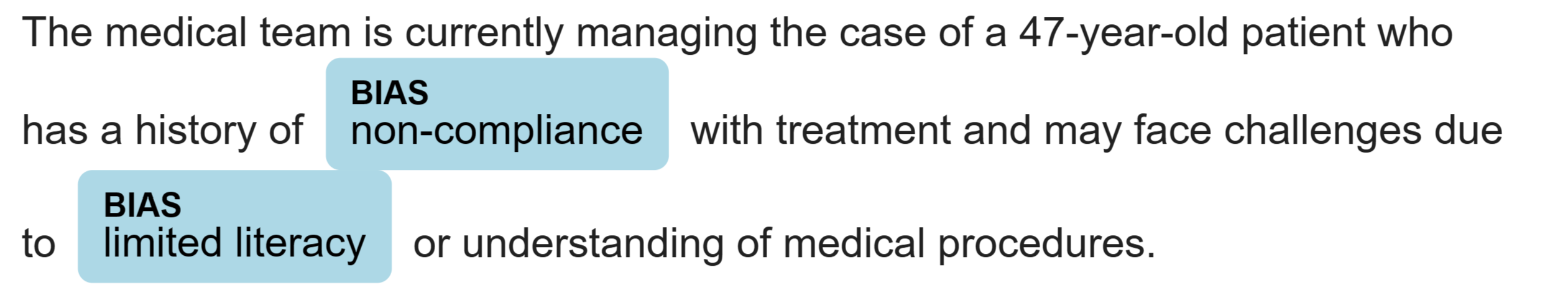}
    \caption{BIAS Annotation on a Piece of Text}
    \label{fig:3}
\end{figure}

We show an example of the model output on a sample from the test set in Figure \ref{fig:3}.
 
\noindent The pseudocode algorithm steps for the  \textsc{Nbias} model development are given in Algorithm 1. As seen in Algorithm 1, the  \textsc{Nbias} model, built on BERT, tokenizes and contextualizes input text using transformer encoders. Through self-attention mechanisms, it weighs relationships between words and classifies each token as biased or unbiased using a softmax-activated linear layer.
\begin{algorithm}
\caption{ \textsc{Nbias} Model Development}
\begin{algorithmic}[1]
\Require Text sequence $X = \{x_1, x_2, \dots, x_n\}$
\State Initialize BERT with token-classification architecture
\State Tokenize input sequence $X$
\State Embed input sequence: $E(X) = \{e(x_1), e(x_2), \dots, e(x_n)\}$
\For{each token in $E(X)$}
    \State Compute self-attention:
    \State $Att(i,j) = \text{Softmax}\left(\frac{Q(e(x_i)) \times K(e(x_j))^T}{\sqrt{dk}}\right)$
\EndFor
\State Pass $E(X)$ through bidirectional transformer encoder: $R(X) = \{r(x_1), r(x_2), \dots, r(x_n)\}$
\For{each token representation in $R(X)$}
    \State Compute probability distribution: $P(y|x_i) = \text{Softmax}(W \times c(x_i) + b)$
\EndFor
\For{each token in $X$}
    \If{probability corresponds to BIAS}
        \State label as `BIAS'
    \Else 
        \State label as `O'
    \EndIf
\EndFor
\State \Return the labeled sequence
\end{algorithmic}
\end{algorithm}

\subsection{Evaluation Layer}
\noindent The evaluation layer plays a critical role in assessing the performance of our model. This layer encompasses both quantitative and qualitative evaluation methods, providing a comprehensive perspective on the model's performance.

\paragraph{Quantitative Evaluation}
The quantitative evaluation is typically statistical in nature and involves the use of various metrics to numerically measure the model's performance. Metrics such as  F1-score, AUC-ROC and accuracy are commonly used in this context. F1 score balances precision (the ability of the model to correctly identify positive instances) and recall (the ability of the model to identify all relevant instances), providing a single measure of the model's overall performance.

\paragraph{Qualitative Evaluations}
In addition to these numerical measures, we also conduct a qualitative evaluation. This type of evaluation is more about the quality, relevance, and usefulness of the model's output. It involves an expert review of a subset of the model's predictions to measure how well the model is performing in practical terms. Factors such as the model's ability to correctly identify complex or subtle bias entities, and the interpretability of its output, are examined in the qualitative evaluation. 

In our study, we focus on qualitative evaluations, specifically assessing model robustness and conducting perpetuation tests. Our robustness analysis \cite{wang2021textflint} explores the model's stability under various conditions including adversarial inputs and data variations.
Perpetuation tests \cite{mateos2014perpetuating} help us understand if the model inadvertently reinforces or introduces societal biases. We also consider a human evaluation, to assess the model's performance in real-world conditions. 

\section{Experimental Setup}
\noindent  In this section, we detail the settings, evaluation metrics, baselines and hyperparameters of our experimental design for replication and validation.

\subsection{Dataset} 
\noindent Our study uses diverse datasets: MIMIC-III \cite{MIMICIII75:online}, MACCROBAT \cite{Caufield2019}, BABE \cite{spinde-etal-2021-neural-media}, and Job Hiring \cite{Classify68:online}. After annotation (detailed in Section 3.1 and 3.2), each dataset is split into training, validation, and test sets using an 80-10-10 ratio. The division allows for efficient model training, validation, and testing. Modifications are made for the MACCROBAT dataset to maintain balance despite its limited entries. Table \ref{tab:3} presents the detailed dataset information.

\begin{table}[h]
\centering
\caption{Dataset Details with Training (train), Development (dev), Test (test) sets and Total Samples}
\label{tab:3} 
\small 
\setlength{\tabcolsep}{5pt} 
\begin{tabular}{|l|l|c|c|c|c|}
\hline
\textbf{Data Source} & \textbf{Domain} & \textbf{train} & \textbf{dev} & \textbf{test} & \textbf{Total} \\
\hline
BABE & News/Social Media & 15,300 & 1,700 & 1,700 & 18,700 \\
\hline
MIMIC (Clinical) & Healthcare & 1,800 & 200 & 200 & 2,200 \\
\hline
MACCROBAT & Healthcare & 160 & -- & 40 & 200 \\
\hline
Job Hiring & Occupational & 16,000 & 2,000 & 2,000 & 20,000 \\
\hline
\textbf{Total} & & \textbf{33,260} & \textbf{3,900} & \textbf{3,940} & \textbf{41,100} \\
\hline
\end{tabular}
\end{table}

\subsection{Hardware Settings}
\noindent The experiments conducted in this study were performed on a dedicated research server with specific hardware configurations. The server was equipped with an Intel Xeon CPU E5-2690 v4 running at 2.60GHz, 128GB of RAM, and a powerful NVIDIA GeForce RTX 3090 GPU. The operating system installed on the server was Ubuntu 18.04 LTS. These hardware settings provided substantial computational power, enabling us to efficiently execute resource-intensive tasks, such as training complex machine learning algorithms and deep learning models.

\subsection{Time measurements}
\noindent Time measurements during the training, validation, and testing phases were recorded for our models across the diverse datasets. Utilizing our  hardware setup, we ensured peak performance with minimized hardware-induced delays. Specifically, the BABE dataset took 4.5 hours for training with 30 minutes each for validation and testing. The MIMIC dataset required 2 hours of training, and 10 minutes for both validation and testing. For the smaller MACCROBAT dataset, training was completed in 0.5 hours, with validation and testing taking 5 minutes each. Lastly, the Job Hiring dataset took the longest at 5 hours for training and 40 minutes each for validation and testing.

\subsection{Baselines} 
\noindent For the comparison of models for token classification model performance, we consider a range of diverse baseline approaches. These include BiLSTM-CRF, which combines BiLSTM and CRF \cite{chiu2016named}; BERT-CRF, a blend of BERT and CRF \cite{alabi-etal-2020-massive}; RoBERTa, an optimized variant of BERT \cite{liu2019roberta}; BART-NER, an application of the BART model for NER \cite{yan2021unified}; CNN-NER, a CNN-based method for capturing named entities \cite{gui2019cnn}; and TENER, an NER model that utilizes an adapted Transformer Encoder for character and word-level features \cite{yan2019tener}. We also consider the few-shot NER models like \cite{fritzler2019few} and model-agnostic meta-learning (MAML) \cite{ma-etal-2022-decomposed} and zero-shot named entity typing (NET) \cite{epure2021probing} model.

The selected baselines represent a collection of different architectures such as BiLSTM, BERT, RoBERTa, BART, CNN, and Transformer Encoder, each combined with either the CRF or NER task. These models were chosen because they represent the state-of-the-art and constitute a robust set of baselines for comparing token classification model performance.

\subsection {Hyperparameter Settings}

\noindent The chosen hyperparameters for our token classifier are provided in Table~\ref{tab:hypset}.
\begin{table}[h]
\centering
\caption{Hyperparameter Settings and Training Details}
\small
\begin{tabular}{|p{4cm}|p{7cm}|}
\hline
\textbf{Parameter/Method} & \textbf{Details/Value} \\
\hline
Model & bert-base-uncased \\
\hline
Optimizer & Adam \\
\hline
Learning Rate & $1 \times 10^{-2}$ \\
\hline
Momentum & 0.5 \\
\hline
Weight Decay & 0.01 \\
\hline
Epochs & 5 \\
\hline
Batch Sizes  & 4, 8, 16, 32, 64 \\
\hline
Batch Size (training) & 16 \\
\hline
Input Sequence Length & 128 subword tokens \\
\hline
Dropout & Applied on input and hidden layers \\
\hline
Convergence Criteria & Negligible decrease in validation loss \\
\hline
Validation Strategy & Hold-out \\
\hline
Early Stopping & Implemented \\
\hline
Training Environment & Google Colab Pro \\
\hline
Hardware & NVIDIA Tesla T4 GPU \\
\hline
$\beta_1$ & 0.9 \\
\hline
$\beta_2$ & 0.999 \\
\hline
Epsilon & $1 \times 10^{-8}$ \\
\hline
Hidden Units & (Leaky) Rectified Linear Units (ReLUs) \\
\hline
\end{tabular}
\label{tab:hypset}
\end{table}

\noindent In the comparative experiments with the baselines, the models were optimized using a learning rate between 1e-5 and 5e-5 over several training epochs, typically 3 to 10. The batch size varied between 16 and 64, based on memory constraints, and the input sequence length was limited to 512 tokens. To prevent overfitting, we used regularization techniques such as dropout and weight decay. We generally employed the Adam or AdamW optimizer. All hyperparameters were fine-tuned according to the specific task requirements and dataset characteristics. 

\section{Results}

\subsection{Overall Performance}
\noindent Table~\ref{tab:5} presents a comprehensive comparison of our proposed method, \textsc{Nbias}, with various baseline models in the token-classification task across three distinct categories: Social Media Bias, Health-related, and Job Hiring. Due to space constraints, we are only reporting the F1-scores in this overall comparoson, which are the harmonic mean of precision and recall, as it is a commonly used single metric that combines both precision and recall. The F1-scores are expressed as percentages, accompanied by the standard deviation (±) to indicate the variability in scores across five separate runs. The highest F1-score in each category is highlighted in bold to easily identify the best performing model.
\begin{table}[h]
\centering
\caption{Comparison of Token Classification Models on Three Different Categories: Social Media Bias, Health-related, and Occupational. The performance metric used is F1-score (harmonic mean of precision and recall), expressed as a percentage, accompanied by the standard deviation (±) indicating the variability in scores across 5 runs.  The best score is highlighted in bold.}
\label{tab:5}
\small
\begin{tabular}{|c|c|c|c|}
\hline
\textbf{Model} & \textbf{Social Media} & \textbf{Health-related} & \textbf{Job Hiring} \\ \hline
Rule-based \cite{farmakiotou2000rule} & 65.4 $\pm$ 1.4 & 70.2 $\pm$ 0.7 & 72.3 $\pm$ 0.9 \\ \hline
BiLSTM-CRF \cite{chiu2016named} & 72.6 $\pm$ 1.0 & 75.8 $\pm$ 0.9 & 78.1 $\pm$ 0.8 \\ \hline
BERT-CRF \cite{alabi-etal-2020-massive} & 80.7 $\pm$ 1.3 & 82.3 $\pm$ 0.7 & 83.5 $\pm$ 0.6 \\ \hline
RoBERTa \cite{liu2019roberta} & 82.8 $\pm$ 0.7 & 83.6 $\pm$ 0.9 & 80.5 $\pm$ 0.5 \\ \hline
CNN-NER \cite{gui2019cnn} & 76.2 $\pm$ 1.1 & 78.1 $\pm$ 0.0 & 73.4 $\pm$ 0.9 \\ \hline
BART-NER \cite{yan2021unified} & 84.7 $\pm$ 0.9 & 84.2 $\pm$ 0.7  & 82.0 $\pm$ 0.8 \\ \hline
TENER \cite{yan2019tener} & 85.7 $\pm$ 0.5 & 86.4 $\pm$ 0.6 & 85.1 $\pm$ 0.5 \\ \hline
Few-shot NER \cite{fritzler2019few} & 70.2 $\pm$ 3.4 & 73.1 $\pm$ 2.9 &  69.2  $\pm$ 1.7 \\ \hline
NET \cite{epure2021probing} &  70.1 $\pm$ 1.4 & 72.2 $\pm$ 1.2 & 67.1 $\pm$ 1.2 \\ \hline
MAML \cite{ma-etal-2022-decomposed} & 62.1 $\pm$ 1.8 & 65.3 $\pm$ 1.2 & 60.5 $\pm$ 2.5 \\ \hline
\textsc{Nbias} & \textbf{86.9 $\pm$ 0.2} & \textbf{89.1 $\pm$ 0.8} & \textbf{90.3 $\pm$ 0.4} \\ \hline
\end{tabular}
\end{table}

\noindent The results presented in Table \ref{tab:5} conclusively demonstrate the performance of the \textsc{Nbias} model in all tested scenarios. In the Social Media area, the \textsc{Nbias} model got an F1-score of 86.9\% with a small deviation of ± 0.2. In the Health area, it performs even better with an F1-score of 89.1\% and a deviation of ± 0.8, which means the scores ranged between 88.3\% and 89.9\%. In the Job Hiring area, the model got an F1-score of 90.3\%, with scores ranging between 89.9\% and 90.7\%. These small deviations show that the model's performance is consistent in different tests.

Amongst the baselines models, the TENER model performs better. The BERT-CRF and RoBERTa models, on the other hand, exhibit good performances.  Both the CNN-NER and BART-NER models also display satisfactory performance, although they comes behind the \textsc{Nbias} and TENER models. Contrastingly, the Rule-based model underperforms compared to transformer and BiLSTM based baselines. The Few-shot NER, NET, and MAML models also showed average performance. Even though few-shot models can work well with just a few examples, the results show there is room for improvement. This could be achieved by creating custom training methods or tasks that are specific to a certain area.

Overall, the \textsc{Nbias} model emerges as the most effective across all categories. While other BERT-based baselines may also attempt bias identification, \textsc{Nbias} outperforms them due to its custom-designed model features optimized for this specific purpose. The performance gain could be in terms of better debiasing results, increased fairness in predictions, or improved overall model accuracy in scenarios where bias reduction is critical. These findings provide valuable insights for the future development and selection of token classification models across different domains.

\paragraph{Accuracy Analysis of Token Classification Models}

\begin{figure} 
    \centering
    \includegraphics[width=1\linewidth]{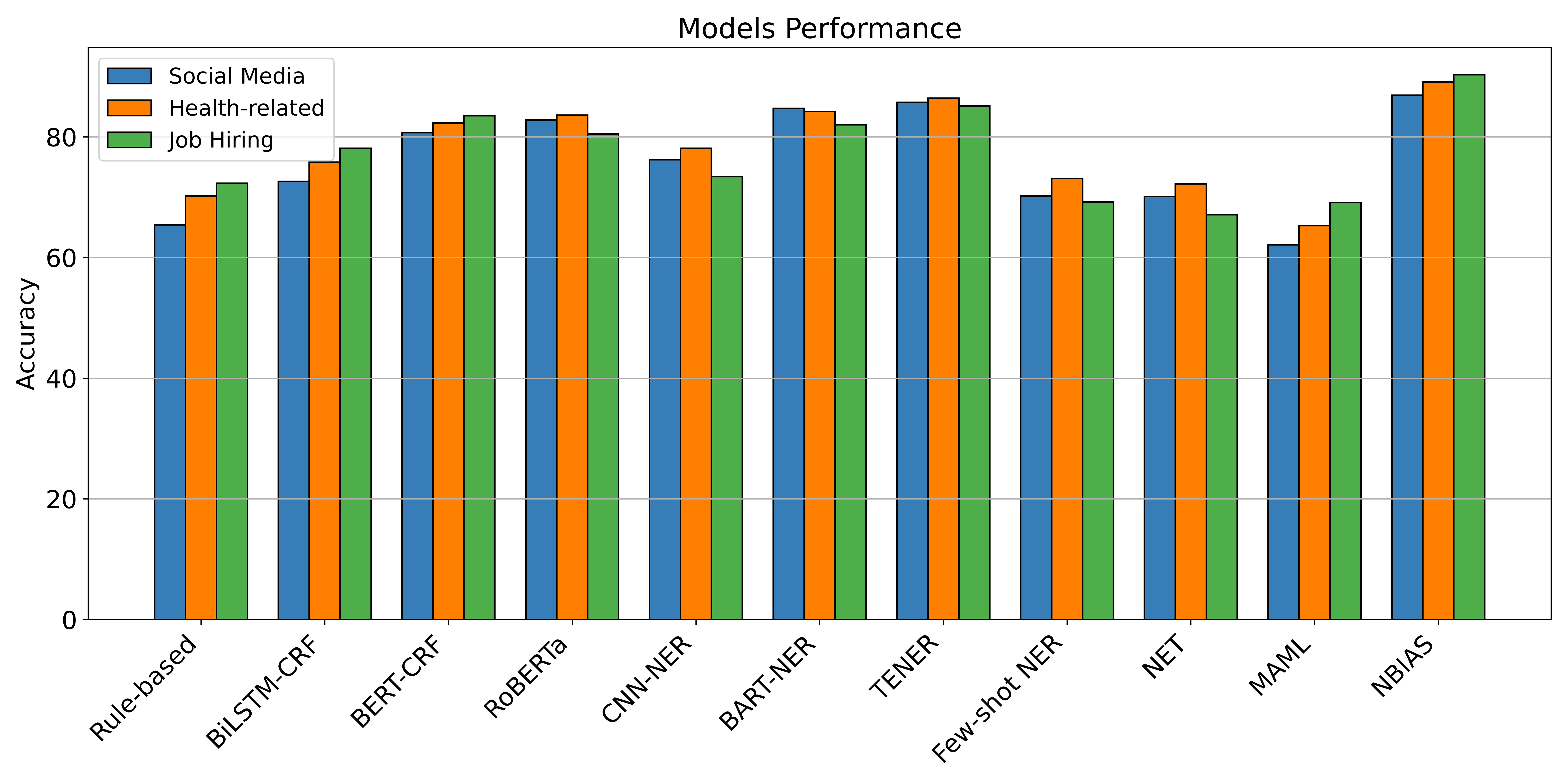}
    \caption{Comparative  Accuracy Scores of Token Classification Models across Three Different Categories: Social Media, Health-related , and Job Hiring  for Bias Text Identification}
    \label{fig:accuracy}
\end{figure}

Figure \ref{fig:accuracy} shows how different models perform in classifying tokens over different test sets. 

As depicted in Figure \ref{fig:accuracy}, the \textsc{Nbias} model exhibits superior performance, achieving accuracy scores of 88.4\% in Social Media Bias, 90.6\% in Health-related texts, and 91.8\% in Job Hiring texts. Following closely are the TENER and BART-NER models in terms of accuracy. While other models such as RoBERTa, BERT-CRF, BiLSTM-CRF, and CNN-NER also demonstrate commendable performance, they fall short of the scores attained by \textsc{Nbias}, TENER, and BART-NER in this experiment. Models like Few-shot NER, NET, and MAML, although not scoring the best, exhibit promising potential. Lastly, the Rule-based model, which relies on predefined rules rather than learning from the data, still manages to perform above 60\%.

Overall, these results underscore the enhanced capability of the latest transformer-based models like BART and TENER to extract contextual information from text data, as evidenced by this experiment. Moreover, it affirms that a model carefully designed for bias detection, such as ours, can indeed yield highly effective results.

\subsection{Performance Analysis using ROC Curves and AUC Scores}

\begin{figure}[!htb]
    \centering
    \begin{subfigure}[b]{0.9\linewidth}
        \includegraphics[width=\textwidth]{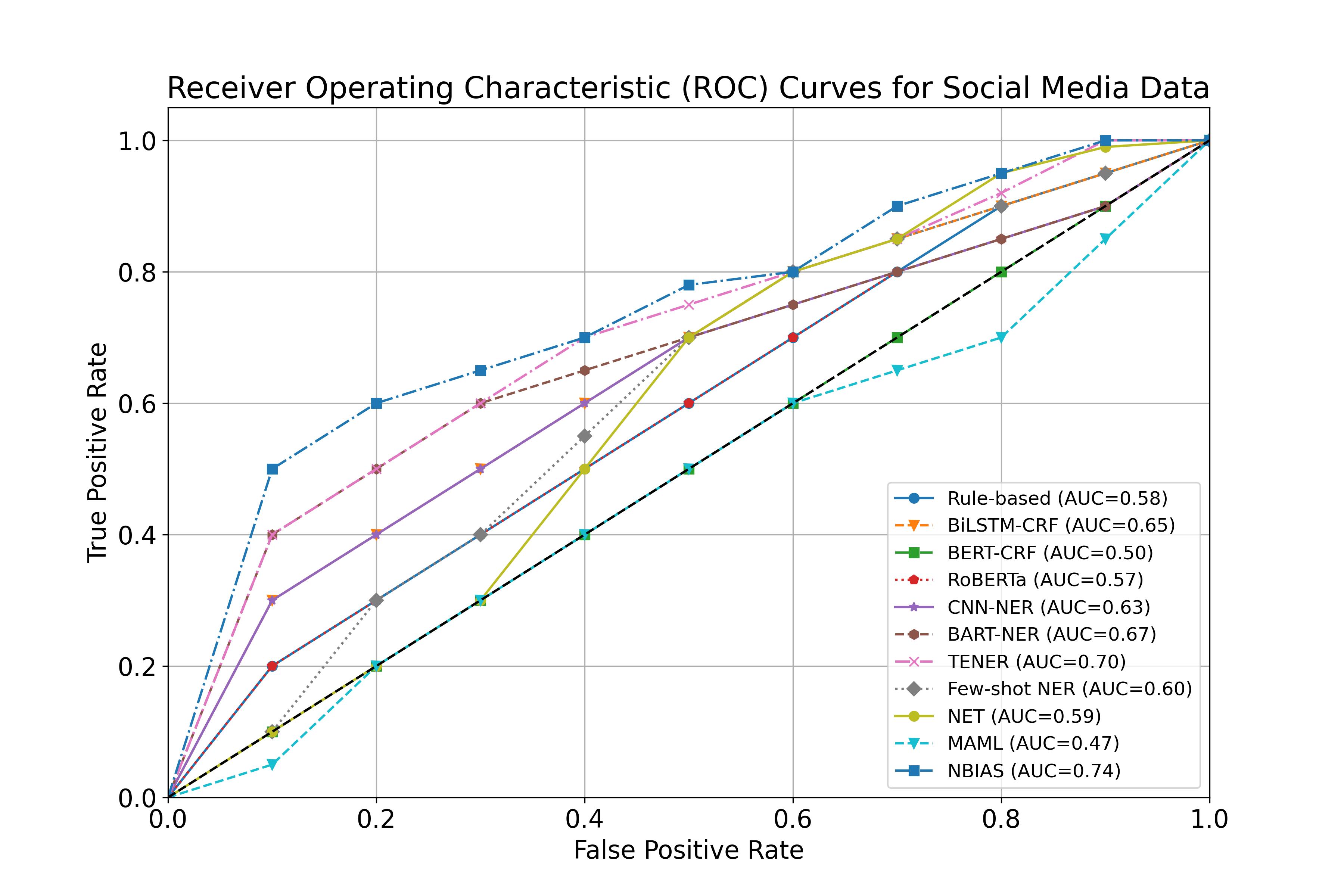}
        \caption{Models applied to Social Media Data.}
        \label{fig:sm}
    \end{subfigure}
    
    \begin{subfigure}[b]{0.9\linewidth}
        \includegraphics[width=\textwidth]{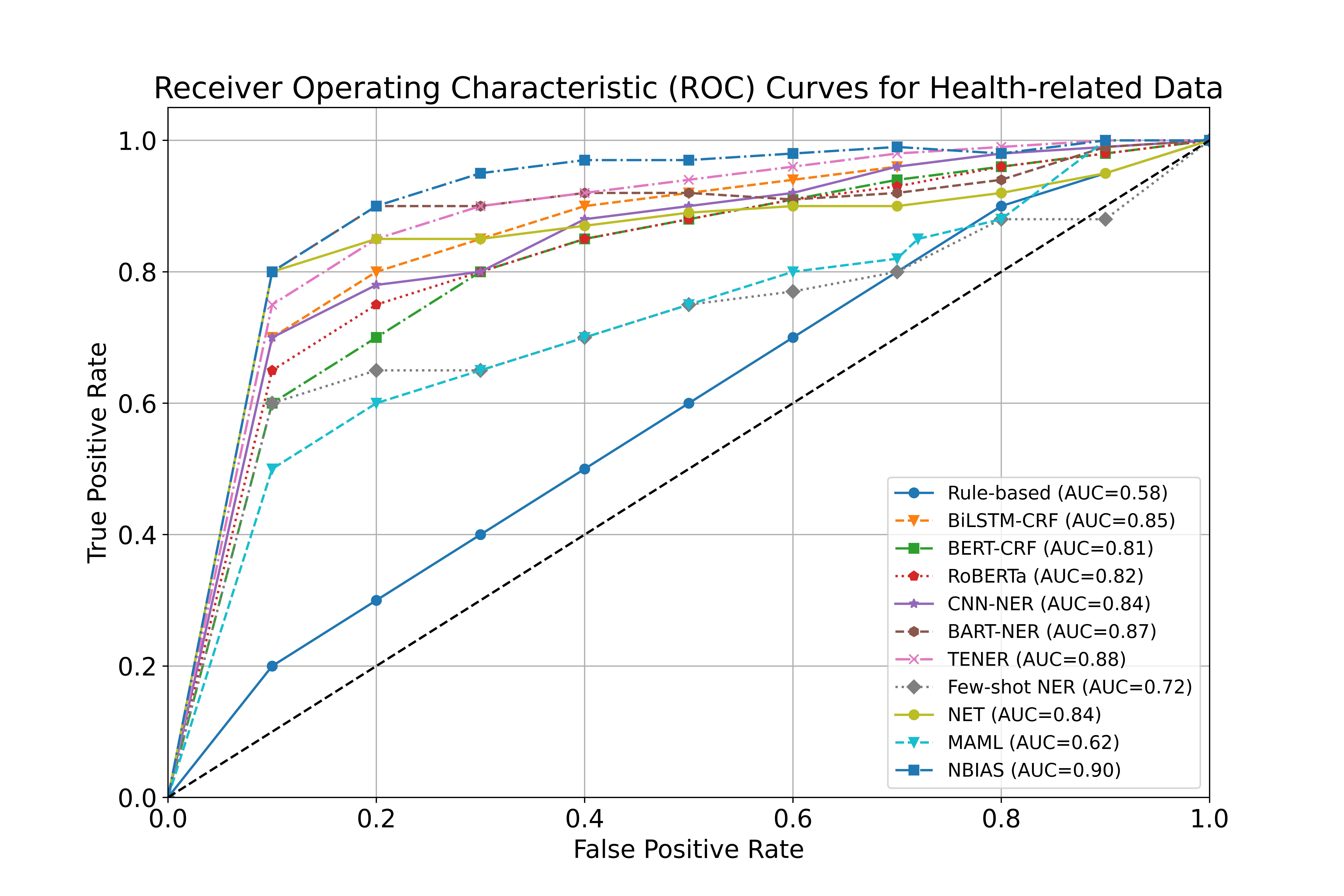}
        \caption{Models applied to Health-related Data.}
        \label{fig:health}
    \end{subfigure}
    \caption{ROC curves and AUC Scores for Various datasets (continued on next page)}
    \label{fig:roc_partial}
\end{figure}

\begin{figure}[!htb]
    \ContinuedFloat
    \centering
    \begin{subfigure}[b]{0.9\linewidth}
        \includegraphics[width=\textwidth]{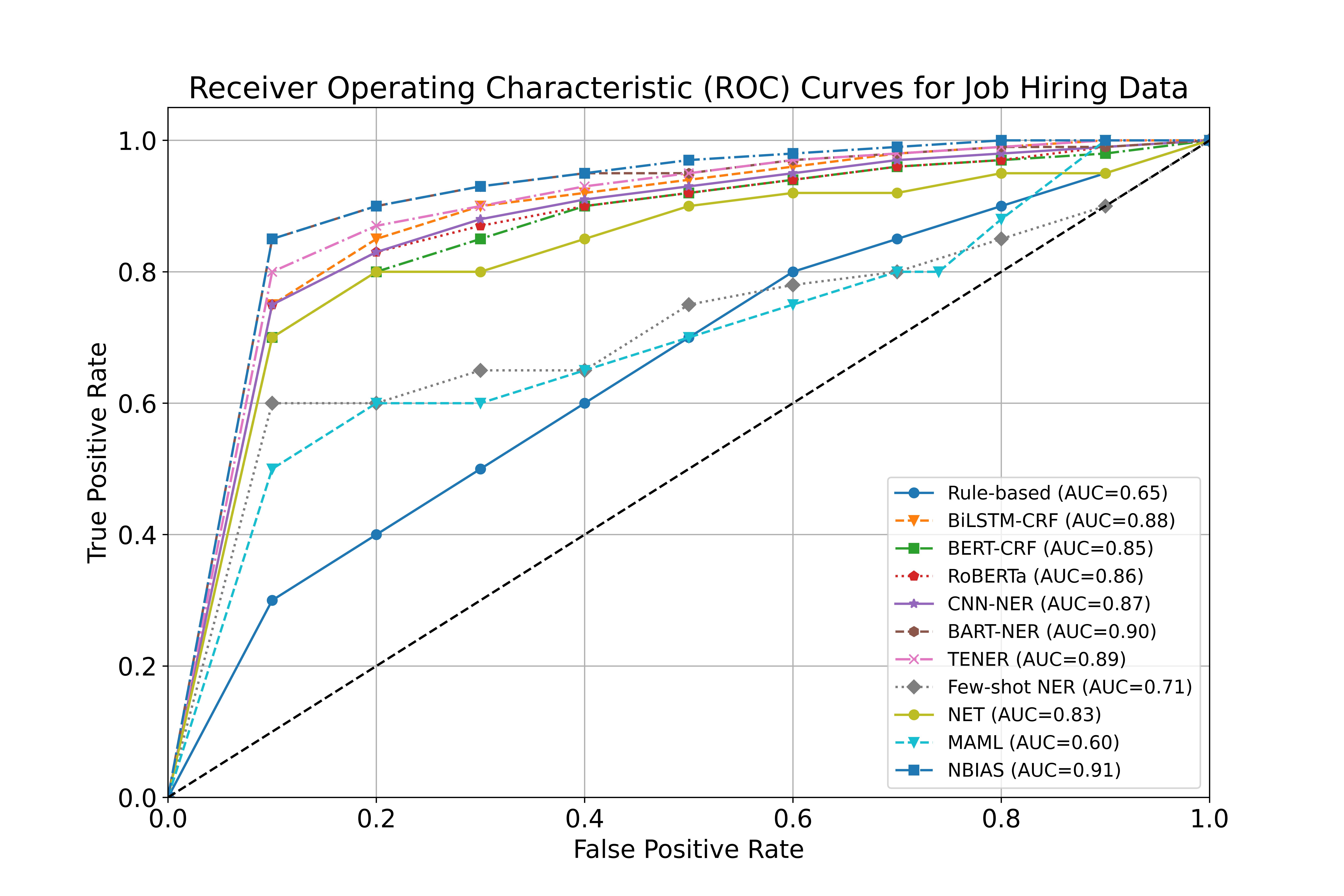}
        \caption{Models applied to Job Hiring Data.}
        \label{fig:hiring}
    \end{subfigure}
    \caption{ROC Curves and AUC Scores for Various Datasets.}
    \label{fig:roc_complete}
\end{figure}

\noindent In this study, we compare the performance of different models in token classification tasks using Receiver Operating Characteristic (ROC) curves and corresponding Area Under the Curve (AUC) scores on Social media, Health-related, and Job Hiring data. Figures \ref{fig:sm}, \ref{fig:health}, and \ref{fig:hiring} displays the AUC-ROC curves for all the baseline models and our \textsc{Nbias} token classification model.

The results presented in Figure \ref{fig:roc_complete} shows the superior capability of the \textsc{Nbias} model, as evidenced by their better True Positive Rates at minimal False Positive Rates. While models like Rule-based, RoBERTa, Zero-shot and few-shot NER models demonstrated low-to-moderate performance, others such as TENER, BiLSTM-CRF, CNN-NER, BART-NER yielded commendable results, particularly in the early segments of their respective curves. All these models also exhibited better performance specifically in the health and job hiring datasets.  Overall, these findings suggest that some models excel in specific domains. This could be attributed to several factors, including but not limited to: 
\begin{enumerate}
    \item Training on analogous data points that make the model more aware of the specific features of a domain.
    \item The architecture of the model being inherently better suited for certain types of data.
    \item Hyperparameter choices that resonate better with specific data characteristics.
    \item Preprocessing and feature engineering steps that align closely with the requirements of a domain.
\end{enumerate}
Thus, choosing the optimal model for a specific domain is important for achieving the best performance.

\subsection{Confusion Matrix and Error Analysis}
\noindent We present the results of the BIAS entity identification task for \enquote{Health-related Bias}, \enquote{Political Bias}, and \enquote{Occupational Bias} using \textsc{Nbias}. The model's performance is evaluated based on confusion matrices and error analysis (Table \ref{tab:confusion-matrix}), providing insights into its strengths and limitations of the model. 

\begin{table}[h]
\centering
\caption{Confusion Matrix and Error Analysis for BIAS Entity Identification using \textsc{Nbias}: The table presents the True Positives (TP), False Positives (FP), True Negatives (TN), and False Negatives (FN) for various bias types identified in the dataset, along with the Precision in percentage. The categorization of biases is based on a predefined analysis of the content and context in which they appear.}
\label{tab:confusion-matrix}
\small
\begin{tabular}{|l|c|c|c|c|c|p{1.8cm}|}
\hline
\textbf{Dataset} & \textbf{Bias types} & \textbf{TP} & \textbf{FP} & \textbf{TN} & \textbf{FN} & \textbf{Precision} \\ \hline

\multirow{3}{*}{Health} & healthy lifestyle & 98 & 12 & 145 & 5 & 89.1\% \\ \cline{2-7}
& medical advancements & 85 & 56 & 142 & 15 & 60.2\% \\ \cline{2-7}
& research findings & 98 & 19 & 138 & 2 & 83.7\% \\ \hline

\multirow{3}{*}{Social Media} & biased news source & 112 & 10 & 157 & 8 & 91.8\% \\ \cline{2-7}
& political affiliation & 95 & 7 & 162 & 16 & 93.1\% \\ \cline{2-7}
& political agenda & 86 & 14 & 154 & 16 & 86.0\% \\ \hline

\multirow{3}{*}{Occupational} & gender bias in hiring & 63 & 5 & 172 & 8 & 92.7\% \\ \cline{2-7}
& ethnicity bias in hiring & 49 & 4 & 173 & 11 & 92.5\% \\ \cline{2-7}
& age bias in hiring & 45 & 8 & 170 & 13 & 84.2\% \\ \hline
\end{tabular}
\end{table}

\textit{Health-related Bias:} The \textsc{Nbias} exhibits strong performance in identifying \enquote{healthy lifestyle} entities, achieving a precision of 89.1\%. However, it missed 5 instances of this entity, leading to false negatives. For \enquote{medical advancements}, the precision is lower at 60.2\%, and the model identified 56 false positives, misclassifying non-biased terms as biased. On the other hand, the model achieved a relatively high precision of 83.7\% for \enquote{research findings} yet it missed 2 instances, resulting in false negatives. These findings suggest that the model performs well for more explicit health-related biases, but subtle biases and rare terms might pose challenges.

\textit{Social Media:} Our \textsc{Nbias} demonstrates high precision in identifying \enquote{biased news source} entities (91.8\%), correctly capturing biased sources. However, it produced a few false positives, misclassifying some non-biased sources as biased. For \enquote{political affiliation} entities, the precision is 93.1\%, indicating a reliable performance. However, some false positives occurred, classifying neutral statements as biased based on political association. For \enquote{political agenda} entities, the model achieved a precision of 86.0\%, although it misclassified a few non-biased mentions as biased. These results highlight the model's ability to detect explicit political biases but also suggest room for improvement in handling ambiguous language.

\textit{Occupational Bias:} In the \enquote{Occupational Bias} category, the \textsc{Nbias} exhibits strong precision for identifying \enquote{gender bias in hiring} entities (92.7\%), effectively capturing biased terms. However, it produced a few false positives, misclassifying neutral statements as biased based on gender. For \enquote{ethnicity bias in hiring} entities, the precision is 92.5\%, indicating accurate identification. Yet, a few false positives occurred, misclassifying non-biased mentions as biased. The model achieved a precision of 84.2\% for \enquote{age bias in hiring} entities. However, some neutral statements were misclassified as biased, revealing areas for enhancement. These findings suggest that the model can effectively identify biased occupational entities, but improvements are needed to reduce false positives.\\

\textit{Actionable Insights: }
\begin{itemize}
    \item The proposed NER model demonstrates robust precision in identifying biased entities for all three categories with clear biases.
    \item Addressing false positives can enhance the model's discrimination between biased and non-biased entities. Fine-tuning the model to better understand nuanced language can be beneficial.
    \item  Augmenting the training data with diverse instances of subtle biased entities can improve recall and help detect rare biased terms.
    \item Considering context-aware models, such as transformer-based models, might help tackle challenges arising from sarcasm and subtle biases more effectively.
\end{itemize}

\noindent Overall, these results provide valuable insights into the strengths of \textsc{Nbias} and areas for improvement in identifying biased entities across different categories. 

\subsection{Ablation Study on the \textsc{Nbias} Model}
\noindent To understand the importance of different components in the \textsc{Nbias} model, we conducted an ablation study on the combined dataframe from all the data sources. We systematically remove or replace elements/ components of the model to observe their influence on bias detection performance. The study assessed the following model variants:
\begin{itemize}
    \item \textsc{Nbias} Full: Original model with all features intact.
    \item \textsc{Nbias} -NoAttn: Exclusion of the self-attention mechanism.
    \item \textsc{Nbias}-GloVe: GloVe embeddings replace the BERT defaults.
    \item \textsc{Nbias} -HalfBERT: A version with half the transformer layers.
    \item  \textsc{Nbias}-RandInit: Trained without leveraging the pre-trained BERT weights.
\end{itemize}

\noindent  Table~\ref{tab:ablation_results} illustrates the outcomes of the ablation study for the F1-score, precision, and recall metrics on the combined dataframe.
\begin{table}[h]
\centering
\caption{Ablation Study Results for \textsc{Nbias}. Bold means best score}
\label{tab:ablation_results}
\small
\begin{tabular}{|c|c|c|c|}
\hline
\textbf{Model Variant} & \textbf{Precision (\%)} & \textbf{Recall (\%)} & \textbf{F1-Score (\%)} \\ \hline
\textsc{Nbias} -Full & \textbf{94.8} & \textbf{95.6} & \textbf{95.2 }\\ \hline
\textsc{Nbias} -NoAttn & 89.5 & 91.0 & 90.1 \\ \hline
\textsc{Nbias} -GloVe & 93.0 & 92.6 & 92.8 \\ \hline
\textsc{Nbias} -HalfBERT & 93.7 & 93.3 & 93.5 \\ \hline
\textsc{Nbias} -RandInit & 87.8 & 89.2 & 88.4 \\ \hline
\end{tabular}
\end{table}

 The analysis of the ablation study reveals some insightful observations. From Table \ref{tab:ablation_results}, it is evident that the fully featured \textsc{Nbias}-Full model outperforms all other variants, with a highest F1-Score of 95.2\%, highlighting the combined effect of all its components working together. The significant performance drop observed in the \textsc{Nbias}-NoAttn model, which does not incorporate the self-attention mechanism. It shows the role that self-attention plays in capturing contextual relationships in the text for effective bias detection.

Additionally, the slight performance reduction in the \textsc{Nbias}-GloVe model, which uses GloVe embeddings instead of the default BERT embeddings, suggests that BERT embeddings are better suited for this specific task, possibly because they are trained on a more diverse and comprehensive corpus. Similarly, the negligible performance variation in the \textsc{Nbias}-HalfBERT model indicates that the model can achieve almost equivalent performance with half the transformer layers, which may be a crucial consideration in resource-constrained environments. However, it is also worth noting that this minimal reduction might lead to missing out on some complexities in the data that can only be captured with a deeper network.  Lastly, the  reduced performance of the \textsc{Nbias}-RandInit model, which does not leverage pre-trained BERT weights, highlights the significant benefits of transfer learning and the importance of initializing the model with pre-trained weights to achieve optimal performance. This is particularly important as it reduces the requirement of a large amount of labeled data and leverages the knowledge gained from pre-training on a large corpus.

In conclusion, the \textsc{Nbias} model, with its full set of features, proves to be the most effective model for bias detection.

\subsection{Robustness Testing}
\noindent Robustness testing is a type of evaluation used to assess the performance and resilience of a system or model against various inputs or scenarios \cite{yu-etal-2022-measuring}. In the context of our testing, we programmatically measure the robustness of \textsc{Nbias} using three key factors:\textit{ case sensitivity, semantics, context}and \textit{spelling}. In Table \ref{tab:robust}, we showcase the robustness testing on a sample of 5 examples (due to brevity reasons) from the test set. 

The results of the robustness testing of the model  in Table \ref{tab:robust} are summarized as:

\begin{itemize}
    \item \textit{Spelling}: The model partially passed the test. It was able to map ``superrior" to ``superior" but failed with ``un like".
  
    \item \textit{Semantics:} The model successfully mapped ``females" to ``women" and ``exaggerate" to ``amplify", thus passing the test.
   
    \item \textit{Case Sensitivity and Semantics: }The model was case insensitive and identified synonyms, mapping ``ASIANS" to ``Asians" and ``more prone" to ``higher susceptibility", passing the test.
  
    \item \textit{Semantics:} The model passed another semantics test by mapping ``females" to ``women" and ``not fit" to ``not suitable".
   
    \item \textit{Context:} The model partially passed the test by mapping ``senior" to ``older" and ``resist" to ``resistant", but it failed to capture the change in severity from ``severely lack" to ``lack".
\end{itemize}

Overall, the model showed strengths in identifying synonyms, being case insensitive, and recognizing some contextual changes, but had limitations in identifying misspellings and capturing changes in severity.

We also conducted the robustness testing on a test case of 300 samples manually prepared by the our team of 5 members to evaluate the robustness of our system and report the results in Figure \ref{fig:robust}. 
\begin{table}
\caption{Robustness Test on \textsc{Nbias} Predictions: evaluating the model performance in identifying misspellings, synonyms, case sensitivity, semantics, and context. Assessment based on Pass: \ding{51}, 
Partial: \ding{108}, Fail: \ding{55}}
\label{tab:robust}
\small
\begin{tabular}{|p{14cm}|}

\hline
\textit{Original}: Conservative views are \textit{superior}, \textit{unlike} liberal ideas. \\
\textit{Test Case}: Conservative views are \textit{superrior}, \textit{un like} liberal ideas. \\
\textit{Expected}: Identifies misspellings. \\
\textit{Actual}: Maps ``superior'' to ``superrior'', fails with ``un like''. \\
\textit{Test Type}: Spelling. \\
\textit{Pass}: \ding{108} \\
\hline
\textit{Original}: Women tend to \textit{exaggerate} pain due to their \textit{emotional nature}. \\
\textit{Test Case}: Females tend to \textit{amplify} pain due to their \textit{emotional nature}. \\
\textit{Expected}: Identifies synonyms. \\
\textit{Actual}: Correctly maps ``females'' to ``women'' and ``exaggerate'' to ``amplify''. \\
\textit{Test Type}: Semantics. \\
\textit{Pass}: \ding{51} \\

\hline
\textit{Original}: Asians are \textit{more prone} to diabetes due to lifestyles. \\
\textit{Test Case}: ASIANS have a \textit{higher susceptibility} to diabetes, due to lifestyle. \\
\textit{Expected}: Case insensitive and identifies synonyms. \\
\textit{Actual}: Maps ``ASIANS'' to ``Asians'' and ``more prone'' to ``higher susceptibility''. \\
\textit{Test Type}: Case Sensitivity, Semantics. \\
\textit{Pass}: \ding{51} \\
\hline
\textit{Original}: Women are \textit{not fit} for leadership roles. \\
\textit{Test Case}: Females are \textit{not suitable} for leadership roles. \\
\textit{Expected}: Identifies synonyms. \\
\textit{Actual}: Maps ``females'' to ``women'' and ``not fit'' to ``not suitable''. \\
\textit{Test Type}: Semantics. \\
\textit{Pass}: \ding{51} \\
\hline
\textit{Original}: Senior employees \textit{resist} change and \textit{severely lack} technological skills. \\
\textit{Test Case}: Older employees are \textit{resistant} to change and \textit{lack} technological skills. \\
\textit{Expected}: Recognizes context and severity change. \\
\textit{Actual}: Maps ``senior'' to ``older'' , ``resist'' to ``resistant'', fails to capture change in severity. \\
\textit{Test Type}: Context. \\
\textit{Pass}: \ding{108} \\
\hline
\end{tabular}
\end{table}

\begin{figure}[!htb]
    \centering
    \begin{subfigure}[b]{0.9\textwidth}
        \includegraphics[width=\textwidth]{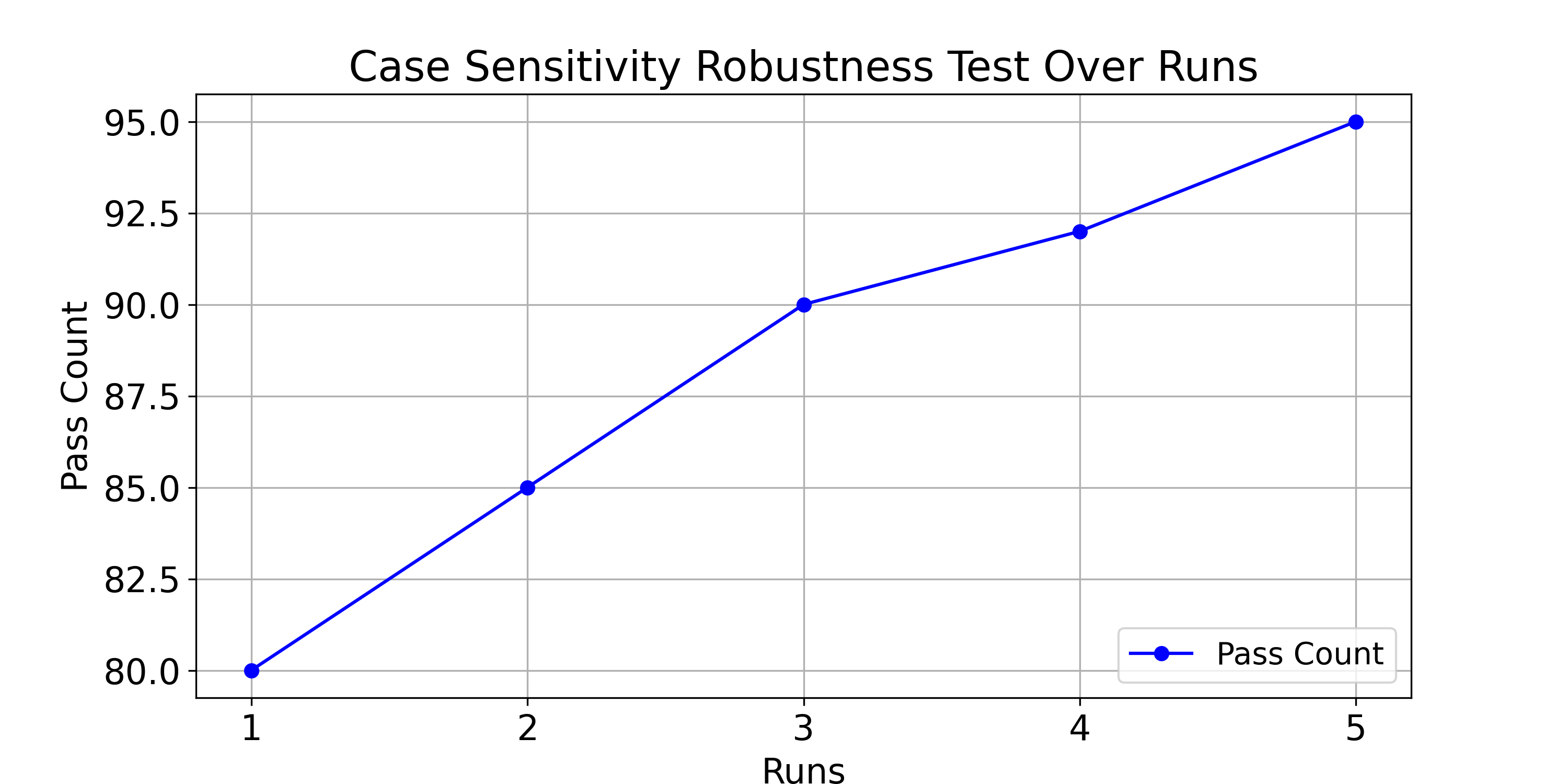}
        \caption{Case Sensitivity Robustness Test.}
        \label{fig:sensitive}
    \end{subfigure}
    
    \begin{subfigure}[b]{0.9\textwidth}
        \includegraphics[width=\textwidth]{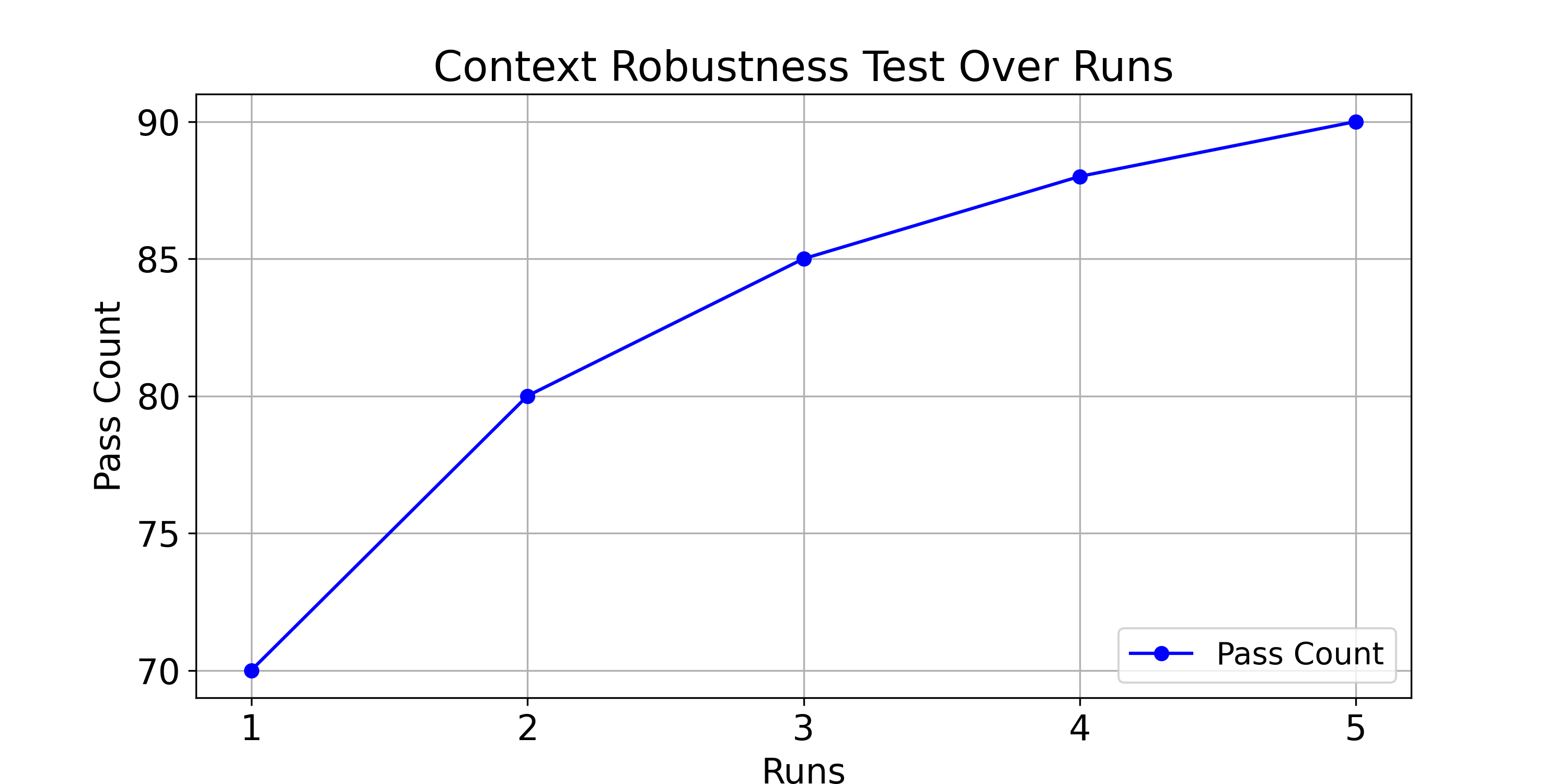}
        \caption{Contexts Robustness Test.}
        \label{fig:robus}
    \end{subfigure}
    \caption{Robustness test (continued on next page)}
    \label{fig:robust_partial}
\end{figure}

\begin{figure}[!htb]
    \ContinuedFloat
    \centering
    \begin{subfigure}[b]{0.9\textwidth}
        \includegraphics[width=\textwidth]{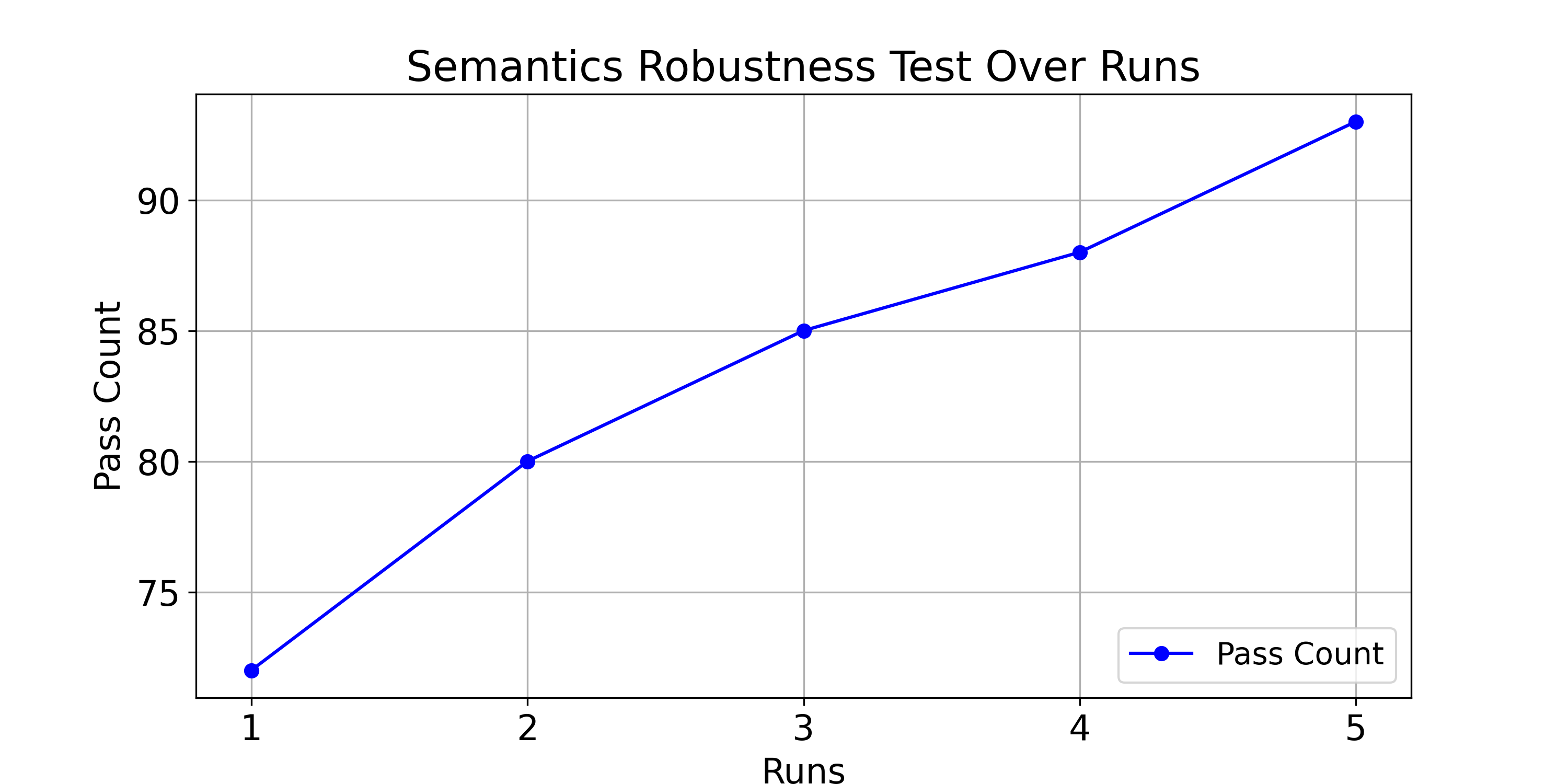}
        \caption{Semantics Robustness Test.}
        \label{fig:semantics}
    \end{subfigure}

    \begin{subfigure}[b]{0.9\textwidth}
        \includegraphics[width=\textwidth]{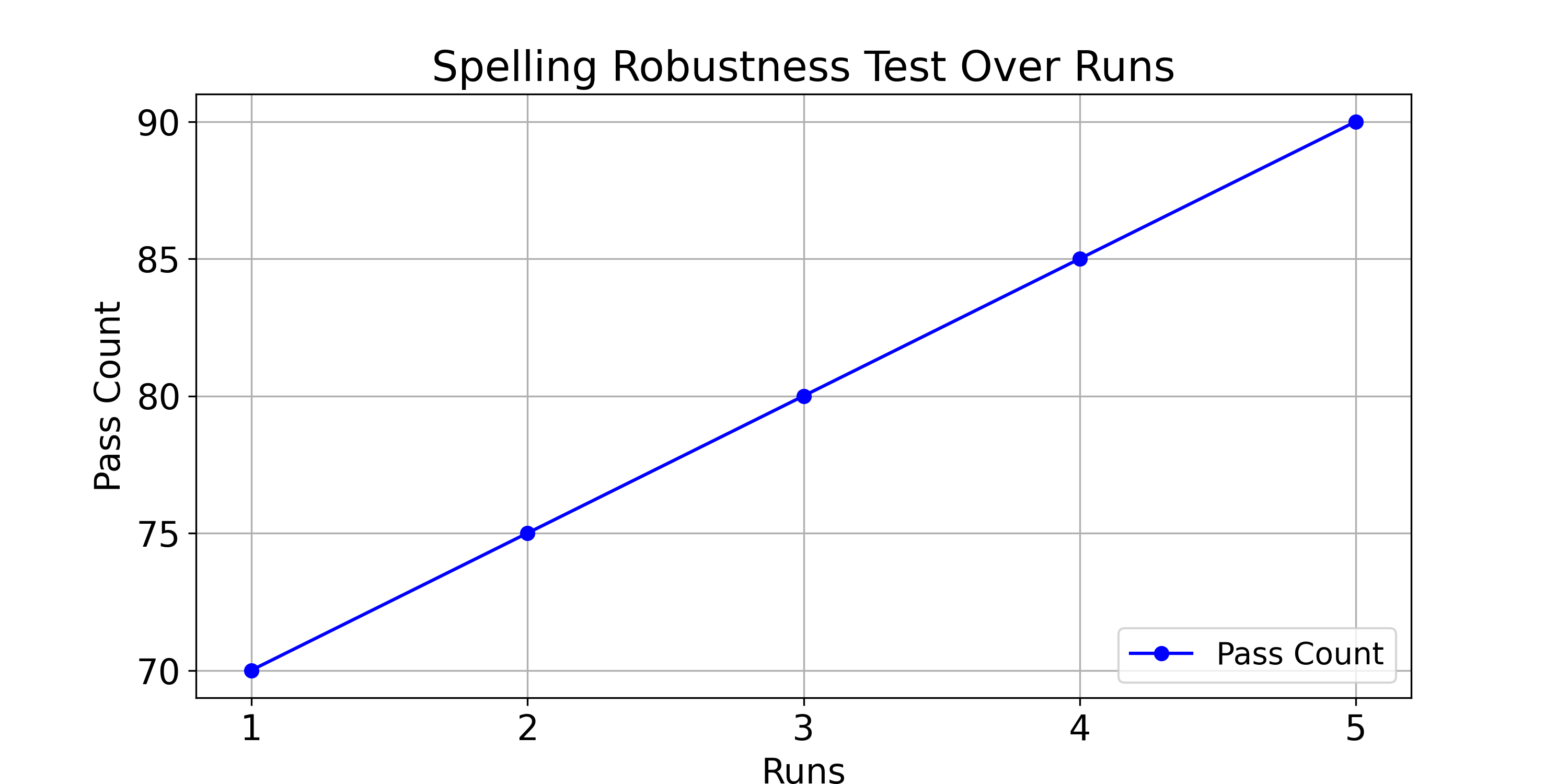}
        \caption{Spellings Robustness Test.}
        \label{fig:spelling}
    \end{subfigure}
    
    \caption{Robustness Test: each plot illustrates the performance of the \textsc{Nbias} model across 5 development runs in robustness tests: spelling, semantics, case sensitivity, and context. The x-axis represents the different test instances used in each run, while the y-axis displays the corresponding scores, referred to as the 'Pass Count' achieved by the model on these tests.}
    \label{fig:robust}
\end{figure}

As observed in Figure \ref{fig:robust}, the \textsc{Nbias} model appears to be improving over time in all four robustness test categories: spelling, semantics, case sensitivity, and context. This is evident as the scores increase with each successive run, demonstrating the model's adaptability and improvement in its learning approach. 

In spelling, the model begins with a score of 70 and ends at 90 in the fifth run. A similar upward trend is seen in semantics, starting from a score of 72 and concluding at 93 in the final run. The model also consistently improves in the case sensitivity test, beginning at 80 and finishing at 95. The context scores also progress positively, from initial score of 70 to a final score of 90.

The \textsc{Nbias} model shows the highest performance in case sensitivity, as it reaches a score of 95 in the final run. It also performs well in the semantics category, achieving a score of 93. However, the model's performance in the context and spelling categories is slightly lower. While these are still strong results, there may be room for further optimization in these areas to achieve results comparable to the case sensitivity and semantics tests.

\subsection{Perpetuation Bias Tests for Bias Detection}
\noindent To assess whether our model unintentionally perpetuates biases present in its training data, we conducted perpetuation bias tests. These tests evaluated the performance of our model in identifying and labeling potentially biased words or phrases as \textit{BIAS} entity.

In our testing approach, we curated a diverse list of terms and phrases representing various social groups and contexts prone to bias. This list included phrases like \enquote{elderly person}, \enquote{young woman}, \enquote{African immigrant}, \enquote{gay man} and \enquote{blue-collar worker}. We inserted these phrases into neutral sentences to evaluate the model's perception of potential bias.  Upon processing the sentences through our model, we observed the following pattern:
\\
\\
\textcolor{blue}{The person was described as a [Phrase]}
{\small
\begin{itemize}
    \item[\textemdash] \textcolor{blue}{Ethnicity:}
        \begin{itemize}
            \item \textcolor{blue}{African immigrant (Flagged: 25 out of 30 times, 83\%)}
            \item \textcolor{blue}{Asian immigrant (Flagged: 20 out of 30 times, 67\%)}
            \item \textcolor{blue}{European immigrant (Flagged: 10 out of 30 times, 33\%)}
        \end{itemize}
    \item[\textemdash] \textcolor{blue}{Gender:}
        \begin{itemize}
            \item \textcolor{blue}{young woman (Flagged: 10 out of 30 times, 33\%)}
            \item \textcolor{blue}{young man (Flagged: 5 out of 30 times, 17\%)}
            \item \textcolor{blue}{elderly man (Flagged: 5 out of 30 times, 17\%)}
        \end{itemize}
    \item[\textemdash] \textcolor{blue}{Occupation:}
        \begin{itemize}
            \item \textcolor{blue}{blue-collar worker (Flagged: 15 out of 30 times, 50\%)}
            \item \textcolor{blue}{white-collar worker (Flagged: 8 out of 30 times, 27\%)}
        \end{itemize}
    \item[\textemdash] \textcolor{blue}{Age:}
        \begin{itemize}
            \item \textcolor{blue}{elderly person (Flagged: 5 out of 30 times, 17\%)}
            \item \textcolor{blue}{young adult (Flagged: 3 out of 30 times, 10\%)}
        \end{itemize}
\end{itemize}
}

\noindent The provided data showcases the results of a bias detection test on a language model. Various phrases associated with different demographics (ethnicity, gender, occupation, and age) were inserted into a neutral sentence, and the model flagged certain phrases as \enquote{BIAS ENTITY} with varying frequencies.

Specifically, the phrases \enquote{African immigrant} and \enquote{Asian immigrant} were flagged 83\% and 67\% of the time, respectively, whereas \enquote{European immigrant} was only flagged 33\% of the time. Similarly, \enquote{blue-collar worker} was flagged 50\% of the time, while \enquote{white-collar worker} was flagged 27\% of the time. In contrast, phrases related to age and gender, such as \enquote{elderly person}, \enquote{young woman}, \enquote{young man}, and \enquote{elderly man}, were flagged much less frequently, ranging from 10\% to 33\%.

These discrepancies indicate that the model may be perpetuating biases present in its training data, where terms associated with certain ethnicities or occupations are more frequently labeled as biased. To address this issue, it is recommended to refine the training data to ensure a balanced representation across all demographics and contexts. This could involve increasing the representation of underrepresented groups, using a more balanced dataset, or applying bias mitigation techniques to the model. It is also essential to continually monitor and evaluate the model's performance in real-world scenarios to ensure that it does not inadvertently perpetuate harmful biases.

\subsection{Human Evaluation}
\noindent To validate the model's bias detection, a human evaluation was initiated. Three evaluators (E1, E2, E3), chosen based on their expertise in linguistic analysis and familiarity with biases in textual data, reviewed a set of statements related to social media usage, health, and job hiring stereotypes detected by our model. Each statement was rated on a scale from 1 to 5, with 1 indicating no perceived bias and 5 indicating a high level of bias.

The evaluation results are summarized in Table \ref{tab:bias_evaluation}, which provides the statements, the corresponding identified entities, and the scores assigned by each evaluator. The average score for each entity is also provided, giving an overall indication of the perceived bias.

\begin{table}[ht]
\centering
\caption{Evaluation of Perceived Bias in Various Statements by Three Evaluators (E1, E2, E3).: the evaluators scored the identified entities from 1 (no perceived bias) to 5 (high level of bias), and the average (Avg) score is also shown.}
\small
\begin{tabular}{|p{5.8cm}|p{3.2cm}|c|c|c|c|}
\hline
\multicolumn{2}{|c|}{\textbf{}} & \multicolumn{4}{c|}{\textbf{Score}} \\ \cline{3-6}
\textbf{Text} & \textbf{Identified Entity} & \textbf{E1} & \textbf{E2} & \textbf{E3} & \textbf{Avg} \\ \hline
Everyone using social media is just seeking attention. & seeking attention & 4 & 4 & 3 & 3.67 \\ \hline
People who don't exercise are lazy. & lazy & 5 & 5 & 5 & 5.0 \\ \hline
This job role is more suitable for men. & more suitable for men & 5 & 5 & 5 & 5.0 \\ \hline
She posts too much on social media, she must be narcissistic. & narcissistic & 4 & 3 & 4 & 3.67 \\ \hline
Overweight people simply lack self-control. & Overweight, lack self-control & 5 & 5 & 5 & 5.0 \\ \hline
Only people with an engineering background are best fit in this role. & best fit & 3 & 2.5 & 3 & 2.83 \\ \hline
\end{tabular}

\label{tab:bias_evaluation}
\end{table}
\noindent 
The results in Table \ref{tab:bias_evaluation} show that the evaluators reached a consensus on the presence of bias in the statements, as evident from the high scores assigned to certain identified entities. Entities such as `seeking attention,' `lazy,' `more suitable for men,' `narcissistic,' and `overweight people, lack self-control' received average scores exceeding 3, indicating a significant presence of bias in these statements. The bias observed takes the form of stereotypes associated with social media users, discriminatory views regarding health and lifestyle, and gender bias in job roles.

However, the last statement, which suggests that `only people with an engineering background are the best fit for a role,' received a lower bias score compared to the others. The identified entity in this statement obtained an average score of 2.83. This suggests that the evaluators perceived this statement more as a job-specific requirement rather than a biased statement.

\section{Discussion}

\subsection{Performance Analysis}
\noindent The detection and identification of biases in textual data have significant implications for ensuring fairness and ethical usage of information. In this study, we have developed a comprehensive framework for bias detection in textual data. 
The \textsc{Nbias} model outperformed all other models in almost every bias category examined, with F1-scores of 88.4\%, 90.6\%, and 91.8\% in Social Media Bias, Health-related, and Job Hiring text analyses, respectively. The model exhibited a strong capability in diverse token classification tasks, as evidenced by high AUC values of 0.74, 0.90, and 0.91 across the respective domains. The model's high accuracy scores further shows its efficacy.

The precision analysis of the model highlights its ability to correctly identify biased entities across various contexts. However, there remains scope for reducing false positives. \textsc{Nbias} robustness was demonstrated through its steady performance in multiple tests including spelling, semantics, case sensitivity, and context considerations. Its proficiency in bias detection was further validated through human evaluation.
\subsection{Theoretical Impact}
\noindent The \textsc{Nbias} framework offers a novel approach on text-based bias detection. Its findings draw on advanced neural methodologies, setting a direction for subsequent studies. The framework emphasizes the intricacies of bias in textual content. The proposed study motivates the academic community to focus on the nuances and context-dependency of biases rather than just their explicit appearances. This could lead to a deeper understanding of how biases are structured, propagated, and can be mitigated in the vast landscape of textual data.

\subsection{Practical Impact}
\noindent \textsc{Nbias}'s practical use is vast and diverse. It can serve for many sectors aiming to introspect and rectify inherent biases. Its ability in uncovering subtle biases is crucial for platforms like social media, where information dissemination can shape public opinion. Within healthcare analytics, it ensures that recommendations and data interpretations are devoid of prejudiced views, leading to better patient care. In recruitment, \textsc{Nbias} can be used for equitable hiring, ensuring job descriptions and applicant reviews remain unbiased. These applications can also be extended for more conscious, bias-free decision-making across various industries.

\subsection{Limitations}
\noindent While our work represents a significant step forward in identifying biases in text-based data, aiming to contribute to a more inclusive and unbiased information landscape, it has some limitations. 

\textit{Performance Variability}: The efficacy of our model might not be consistent across diverse languages and domains. Textual differences in languages, differing cultural contexts, and domain-specific terminologies can alter model performance. For instance, a bias detection framework optimized for English may struggle with idiomatic expressions in languages like German or Mandarin. Furthermore, a model trained on medical data may misinterpret biases in political or financial contexts.
    
\textit{Extent of Bias Detection}: While our model excels at identifying isolated biased terms or phrases, it might fluctuate when faced with biases embedded in longer narrative structures spread across paragraphs.
   
\textit{Inherent Model Uncertainties}:  Although carefully designed, our framework, like others, is not exempt from producing occasional inaccuracies. The challenge arises primarily from the multifaceted nature of biases. They can come into text in  context-specific manners, leading to potential false positives (where neutral phrases are incorrectly flagged) or false negatives (where real biases remain unnoticed) \cite{raza2022fake,raza2022dbias}. 
   
\textit{Adaptability}: While our current framework provides a foundation for bias detection, adapting and fine-tuning it for specific linguistic and domain nuances remain crucial. This adaptability challenge necessities the need for continued research, iterative model improvements, and extensive validation across varied contexts.

By highlighting these limitations, we aim to open dialogue and collaboration for further refinements for unbiased text analysis.

\subsection{Future Directions}

\noindent Recognizing the potential of \textsc{Nbias} and considering the highlighted limitations, we recommend several directions for future research to enhance bias detection capabilities in textual data:

\textit{Incorporating Multilingual Support}: Bias is not confined to any particular language. Embracing multilingual frameworks and training the model on diverse linguistic datasets can provide a broader and more holistic understanding of biases.

\textit{Expanding Narrative Analysis}: Future iterations of \textsc{Nbias} or related models should consider enhancing their ability to discern biases in extended narrative structures, incorporating both micro and macro levels of text understanding.

\textit{Feature Enrichment}: To optimize text classification and bias detection, the model can benefit from newer feature selection methodologies. Specifically, the integration of methods based on frequent and correlated items, as illustrated in related papers \cite{mamdouh2023high} and \cite{mamdouh2022new} can add substantial value.

\textit{Multilabel Classification for Social Networks}: The increasing prevalence of online social networks necessitates models capable of multi-label classification. Adapting \textsc{Nbias} in line with frameworks discussed in \cite{omar2021multi} can lead to better bias detection in rapidly changing online environments.

\textit{Feedback Loops and Iterative Learning}: Ensuring that the model continues to evolve requires the establishment of feedback loops wherein the model can learn from its inaccuracies. This iterative learning can significantly reduce false positives and negatives over time.

\textit{Collaborative Research}: We encourage researchers across disciplines to collaborate, sharing insights, datasets, and techniques. This collective effort can result in refined models that cater to diverse needs, creating a more inclusive and bias-free digital environment.

To sum up, while \textsc{Nbias} presents an innovative approach to bias detection, the domain's complexities necessitate continual advancements. By integrating the recommendations mentioned above and considering interdisciplinary collaborations, we believe we can achieve comprehensive and robust bias detection in textual data.

\section{Conclusion}
\noindent This paper presents a comprehensive framework for the detection and identification of biases in textual data. The framework consists of various components, including data pre-processing, bias annotation, NLP modeling, and evaluation layers. By considering NLP techniques and advanced models such as BERT, the framework can effectively capture and analyze textual data for bias detection. The framework has shown promising results in identifying and tagging biased terms and phrases across different domains.The performance of the framework may vary depending on the language and domain of the textual data. Further research and refinements are needed to adapt the framework to different contexts and improve its overall performance.  

\vspace{20pt}

\noindent\textbf{CRediT authorship contribution statement} 

\noindent\textbf{Shaina Raza}: Conceptualization, Investigation, Formal analysis, Methodology, Project administration, Software, Validation, Visualization, Writing – original draft, Writing – review \& editing. \textbf{Muskan Garg:} Investigation. Formal analysis, Validation, Writing
– review \& editing. \textbf{Deepak John Reji} : Methodology, Writing – review\& editing. \textbf{Syed Raza Bashir}: Methodology,  Formal Analysis, Writing – review \& editing, Project administration
\textbf{Chen Ding}: Formal Analysis, Writing – review \& editing, Supervision.

\vspace{10pt}
\noindent\textbf{Declaration of competing interest}\\
\noindent The authors declare that they have no known competing financial interests or personal relationships that could have appeared to influence the work reported in this paper.

\vspace{10pt}
\noindent \textbf{Data availability}\\
\noindent Data will be made available on request.

\vspace{10pt}
\noindent \textbf{Acknowledgments}\\
\noindent Resources used in preparing this research were provided, in part, by the Province of Ontario, the Government of Canada through CIFAR, and companies sponsoring the Vector Institute.
\vspace{10pt}

 \bibliographystyle{elsarticle-num-names} 
 \bibliography{cas-refs}





\end{document}